%% file: arxivADOTA.tex
\documentclass[twocolumn, 10pt]{IEEEtran}
\usepackage{graphicx}
\usepackage{amssymb}
\usepackage{amsmath}
\usepackage{mathtools}
\usepackage{dsfont}
\usepackage{cite}
\usepackage{stfloats}
\usepackage{psfrag}
\usepackage[mathscr]{euscript}

\usepackage{acronym}  
\usepackage{booktabs}
\usepackage{bbm}
\usepackage{subcaption}
\usepackage{float}
\usepackage{booktabs}
\usepackage{float}
\usepackage{latexsym}
\usepackage{algorithm}

\usepackage{amsmath,amsthm,amssymb}
\usepackage{xcolor}
\usepackage{algorithmicx}
\usepackage{algpseudocode}
\usepackage[noend]{algcompatible}

\algnewcommand\algorithmicreturn{\textbf{return}}
\algnewcommand\RETURN{\State \algorithmicreturn}%

\algrenewcommand\algorithmicrequire{\textbf{function}} 
\algnewcommand\algorithmicreq{\textbf{Require:} }
\algnewcommand\REQ{\STATEx \algorithmicreq{}}%

\usepackage{float}
\usepackage{balance}
\input{aconym}
\input{defmetric}

\newcommand{\paperTitle}{ Adaptive Federated Learning Over the Air }

\begin{document}

{
\title{\paperTitle}

\author{
    
    Chenhao~Wang, \textit{Student Member, IEEE},
    Zihan~Chen, \textit{Member, IEEE},
    Nikolaos Pappas, \textit{Senior Member, IEEE},\\
    Howard H. Yang, \textit{Member, IEEE},
    Tony Q. S. Quek, \textit{Fellow, IEEE}, 
    and H. Vincent Poor, \textit{Life Fellow, IEEE}

\thanks{
     This work was supported in part by the National Natural Science Foundation of China under Grant 62201504, the Zhejiang Provincial Natural Science Foundation of China under Grant LGJ22F010001, the Zhejiang – Singapore Innovation and AI Joint Research Lab, and the Zhejiang University/University of Illinois Urbana-Champaign Institute Starting Fund. The work of N. Pappas was supported in part by the Swedish Research Council (VR), ELLIIT, the European Union (ETHER) under Grant 101096526; in part by the European Union’s Horizon Europe Research and Innovation Programme under the Marie Skłodowska-Curie Grant Project SOVEREIGN under Agreement 101131481; and in part by the HORIZON-MSCA-2022-DN-01 Project ELIXIRION under Grant 101120135. An earlier version of this article was presented at the IEEE International Workshop on Signal Processing Advances in Wireless Communications \cite{ours}.
     (\emph{Corresponding Author: Howard H. Yang})

    C. Wang and H. H. Yang are with the ZJU-UIUC Institute, Zhejiang University, Haining 314400, China (email: chenhao.22@intl.zju.edu.cn, haoyang@intl.zju.edu.cn).

    Z. Chen and T.~Q.~S.~Quek are with the Information Systems Technology and Design Pillar, Singapore University of Technology and Design, Singapore (e-mail: zihan\_chen@sutd.edu.sg, tonyquek@sutd.edu.sg).

    N. Pappas is with the Department of Computer and Information Science, Linköping University, Linköping 58183, Sweden (e-mail: nikolaos.pappas@liu.se).

    H.~V.~Poor is with the Department of Electrical and Computer Engineering, Princeton University, Princeton, NJ 08544 USA (e-mail: poor@princeton.edu).
}
}
\maketitle
\acresetall
\thispagestyle{empty}
\begin{abstract}
We propose a federated version of adaptive gradient methods, particularly AdaGrad and Adam, within the framework of over-the-air model training.
This approach capitalizes on the inherent superposition property of wireless channels, facilitating fast and scalable parameter aggregation. 
Meanwhile, it enhances the robustness of the model training process by dynamically adjusting the stepsize in accordance with the global gradient update.
We derive the convergence rate of the training algorithms, encompassing the effects of channel fading and interference, for a broad spectrum of nonconvex loss functions. 
Our analysis shows that the AdaGrad-based algorithm converges to a stationary point at the rate of $\mathcal{O}( \ln{(T)} /{ T^{ 1 - \frac{1}{\alpha} } } )$, where $\alpha$ represents the tail index of the electromagnetic interference.
This result indicates that the level of heavy-tailedness in interference distribution plays a crucial role in the training efficiency: the heavier the tail, the slower the algorithm converges. 
In contrast, an Adam-like algorithm converges at the $\mathcal{O}( 1/T )$ rate, demonstrating its advantage in expediting the model training process.
We conduct extensive experiments that corroborate our theoretical findings and affirm the practical efficacy of our proposed federated adaptive gradient methods. 
\end{abstract}
\begin{IEEEkeywords}
Federated learning, adaptive gradient method, over-the-air computing, heavy-tailed noise, convergence rate.
\end{IEEEkeywords}
\acresetall

\section{Introduction}\label{sec:intro}
\subsection{Motivation}
Federated learning (FL) is an emerging distributed machine learning paradigm that helps preserve privacy in model training~\cite{MaMMooRam:16,li2020flsurvey,ZhaFenYan:20}. 
A typical FL system consists of an edge server and a group of end-user devices (a.k.a. clients), with each client keeping its data. 
These agents collaborate to train a global model by optimizing a loss function composed jointly by all participants. 
Generally, each round of model training comprises three stages: (intermediate) parameter uploading from the clients, parameter aggregation and model update at the edge server, and broadcasting the updated results from the server to the clients for a new round of local training.
This approach not only fully utilizes the processing power of end-user devices but also addresses the data silo problem arising from the isolation of data among different clients. 
While helping ensure data privacy, FL enables end-users to access a globally applicable model, resolving the constraints of training machine learning models on edge devices. 
Due to this salient advantage, FL has attracted increasing attention in academia and industry, showing considerable potential in various applications, ranging from finance and connected vehicles to smart homes and intelligent healthcare. 

The training process of FL requires frequent parameter exchanges between the clients and the edge server, incurring significant communication overhead \cite{NikDhi:20ComMag}. 
For networks with limited communication resources, such communication bottleneck often strains the training efficiency and inhibits the scalability of the FL system \cite{yang_tcom_fl_scheduling}. 
One possible way to cope with this issue is by integrating analog over-the-air (A-OTA) computations into the FL system, exploiting the superposition property of radio waveforms to promote fast and scalable parameter aggregation~\cite{AmiGun20:TSP,GuoZhu:21JCIN,chen2023ota_mag,SahYanICST}. 
Under this framework, every client constitutes an analog signal composed of a set of common shaping waveforms, each modulated by one element in the gradient vector, and simultaneously transmits it to the edge server.
By filtering out the received signal, the edge server obtains an automatically aggregated gradient --- albeit one that could be significantly distorted --- to update the global model. 
Then, the edge server sends the updated results back to the clients so they can train their local models further.

The A-OTA FL paradigm simultaneously processes model parameters during transmissions, improving spectral efficiency and substantially reducing access latency and energy consumption to edge learning systems~\cite{GuoZhu:21JCIN,ZhuWang:20TWC}.
Despite these properties, the automatic gradient aggregation in OTA FL is achieved at the price of distorting the received signals~\cite{ZhaZhu:22TWC,YangChen:22JSTSP,ZhuXu:21WC,SerCoh:20TSP}. 
More precisely, the random channel fading generally attenuates the radio signal magnitude, deteriorating the precision of the aggregated global gradient. 
Moreover, this is exacerbated by the interference, which usually follows a heavy-tailed distribution, inevitably introduced by the shared nature of wireless channels. 
As a result, the noisy aggregated gradient would cause abrupt direction changes in the model training trajectory, impeding convergence rate and inflicting unstable training performance. 
In conventional machine learning settings, adaptive gradient methods (where the most successful examples are AdaGrad~\cite{ADAGRAD2011} and Adam~\cite{Adam2015}) have cemented their success in robustifying model training~\cite{reddi2020adaptive,ward2020adagradstep,mehta2019cnn}. 
However, whether those results apply to an OTA FL system still needs to be determined.
Therefore, the central thrust of the present article is to close this research gap by developing a systematic scheme to integrate adaptive gradient methods into the A-OTA FL framework and reveal the method's efficacy via rigorous analysis. 

\subsection{Main Contributions}
Adaptive gradient methods leverage the information of all previous gradients observed along the model training process to update the stepsize on the fly, achieving more robust training performance.
Taking AdaGrad as an illustrative example, it accumulates a sum of the squares of all the gradients received up to the current iteration and divides the latest gradient by the square root of the (square) gradient sum to furnish an automatic stepsize schedule.
In the context of A-OTA FL, however, the efficacy of such an approach may need to be revised. 
Due to channel disturbances in the analog transmissions, the global gradient vector is distorted by channel fading (which has a multiplicative effect on the entries of each client's uploaded gradient) and interference (which has an additive effect on the sum of the received gradients).
Since the electromagnetic interference usually obeys a heavy-tailed distribution (e.g., an $\alpha$-stable distribution) \cite{ClaPed:21LCOMM,Mid:1977TEMC,WinPin:09JPROC}, time-average operations such as summing the historical gradients may not be effective in reducing the noise.

It would also not alleviate the multiplicative effects stemming from the channel fading.
To that end, whether a direct extension of the AdaGrad-like method could be suitable for A-OTA FL model training is unclear.
Whether dividing the currently obtained noisy global gradients by a sum of previously accumulated noisy gradients would suppress or amplify those randomness effects remains unknown. 
More broadly, would adaptive gradient methods ever work in an A-OTA FL setting?

This paper responds to this question with an affirmative answer.
The principal contributions of this work are summarized below.
\begin{itemize}
    \item We develop a systematic approach to incorporate adaptive gradient methods into the A-OTA framework. The scheme leverages historical knowledge of the global iterations to perform more informed adjustments in the stepsize, combating impacts of channel fading and interference on model training. 
    Moreover, the proposed method has a low computational complexity and can be implemented in practice. 

    \item We derive convergence rates of our algorithms for non-convex loss functions. The analysis quantifies the effects of various factors on the convergence speed, such as the number of clients, channel fading, and interference. It also characterizes how the hyperparameters in the algorithms influence the convergence performance of the system. 
    Our result reveals that heavy-tailedness in the interference distribution significantly affects the convergence rate of AdaGrad-based algorithms. At the same time, Adam-like model training approaches are more resilient to channel distortions. 
        
    \item We carry out extensive experiments to examine the efficacy of our proposed method. Specifically, we conduct learning tasks of ResNet-18 and ResNet-34 on the EMNIST, CIFAR-10, and CIFAR-100 datasets under different system configurations. Across all the scenarios inspected, our algorithms consistently outperform the traditional FedAvgM method in terms of convergence rate and prediction accuracy, validating its effectiveness in improving system performance. 
    The experimental results also verify our theoretical analysis of the convergence rates for the two proposed adaptive algorithms, showing that the Adam-like algorithm attains a significantly faster convergence rate than the AdaGrad-like method.
\end{itemize}

\subsection{Notation}
To represent scalars and vectors, we use lowercase letters and their bold versions, e.g., $x$ and $\boldsymbol{x}$.
Given a vector $\boldsymbol{x}$, we use $\boldsymbol{x}^\top$ to denote its transpose and $\| \boldsymbol{x} \|_p$, where $p \geq 1$, to denote its \textit{L}-$p$ norm (when $p=2$, it is the normally used Euclidean norm).
Moreover, we adopt $\sqrt[p]{x}$ to represent the $p$-th root of $x$, whereas $\sqrt[p]{\boldsymbol{x}}$ stands for the $p$-th root of $\boldsymbol{x}$ in an entry-wise manner. 
Given two vectors $\boldsymbol{x}$ and $\boldsymbol{y}$, we use $\langle \boldsymbol{x}, \boldsymbol{y}\rangle$ and $\boldsymbol{x}^\top\boldsymbol{y}$ interchangeably to represent their inner product.
For representations related to sets, we use $\{1, 2, \ldots, M\}$ to represent the set containing all integers from 1 to $M$. Also, $\{a_n\}$ means a sequence $a_0, a_1, \ldots, a_n$. 
If $\mathscr{S}$ is a set, its cardinality is denoted by $|\mathscr{S}|$, and $\{ s_n \}_{ n=1 }^N$ indicates a set with elements $s_n$ ranging from $n=1$ to $n=N$.
In the context of function operations, given a function $f: \mathbb{R}^d \rightarrow \mathbb{R}$, we denote by $\nabla f$ its gradient and $\nabla_i f$ the $i$-th component of the gradient.

\section{Related Work}\label{sec:relat}
OTA computations \cite{NazGas:07TIT,GoldBoc:13TSP} refer to a class of schemes that calculate (or approximate) a function of data distributed in multiple end-user devices of a wireless network without reconstructing the data in its entirety. 
Such methods fuse the signal processing and wireless transmission, 
exploiting the superposition property of multiple access channels to realize linear or even nonlinear computational operations on the data transmitted from different sources.
It resolves the issues of spectrum availability when the system confronts a massive number of connected devices.
Recognizing that FL model training only requires computing the sum of clients' uploaded parameters, rather than requesting the precise value of each client's parameter, a line of recent studies \cite{ZhuXu:21WC,AmiGun20:TSP,SerCoh:20TSP,YangChen:22JSTSP,ChenLi:23ICASSP,Sahin:23TWC,YanCheQue:22TSP} proposed integrating OTA computations with FL, arriving at a low-latency multi-access edge learning scheme (which is commonly known as the OTA FL).
Besides reducing delay in the radio access process, OTA FL also features other advantages, including high spectral efficiency, low energy consumption, significantly enhanced system scalability, and (potentially) elevated generalization power~\cite{YangChen:22JSTSP}. 
However, the implementation of OTA FL relies on analog transmissions, which inevitably distorts the received signal by 
introducing channel impairments such as fading and interference to it. 

In response, a few works \cite{YangJia:20TWC, ZhuHua19:IOTJ, ChenZhao:18IOTJ, KimSwin:23TWC} have suggested improving the signal quality in A-OTA FL systems through beamforming designs. 
Specifically, \cite{YangJia:20TWC} demonstrated that devising appropriate beamforming can accelerate the convergence rate of OTA FL systems.
While \cite{ZhuHua19:IOTJ,ChenZhao:18IOTJ} showed that the broadly used zero-forcing approach can optimize communication performance by canceling out cross-talk among transmitters, \cite{KimSwin:23TWC} developed a low-complexity algorithm based on the projected subgradient method, delivering notable improvements via the use of multiple antennae.
Moreover, aided by an adequate estimation of the channel state information (CSI), the training efficiency can be enhanced by adapting the receiver beamforming (i.e., filtering) strategy to the channel variations. 
Indeed, when CSI is available, one can employ power control methods to counteract errors during the aggregation process and improve the performance of OTA FL systems~\cite{ZhaTao:21TWC,CaoZhu:22JSAC,YanQiu:22ISIT}.
Additionally, second-order methods~\cite{KroElg:22TGCN,YanJian:22TWC}, such as the Newton method, can be integrated with the OTA FL training to accelerate the convergence. 

On a separate track, adaptive gradient methods play a crucial role in the model training process of numerous machine learning algorithms. Particularly, AdaGrad~\cite{ADAGRAD2011} stands as a variant of the gradient descent method that re-adjust the step sizes of each coordinate by the sum of squared past gradient values, marking its effectiveness in many instances (but may exhibit suboptimal performance occasionally). Subsequently, RMSprop~\cite{rmsprop} was proposed to address the algorithmic instability issues by employing exponential moving averages instead of cumulative sums. Based on these advances, \cite{Adam2015} proposed Adam, a method that prevails in various model training schemes nowadays.

This paper aims to straddle two representative adaptive optimizers, AdaGrad and Adam, to the OTA FL framework. 
Building upon our previous result \cite{ours}, which only considers the AdaGrad method and Gaussian noise, this paper investigates the more sophisticated Adam scheme and considers the more general heavy-tailed interference.
It is also noteworthy that, in principle, any adaptive optimizer can be employed within our framework.
Indeed, numerous other adaptive methods can also be regarded as specific instances within the framework we propose, with no inherent distinction in their fundamental conceptual underpinnings. Our system is amenable to straightforward extension to encompass these aspects. 

\begin{figure*}[t!]
  \centering{}

    {\includegraphics[width=2\columnwidth]{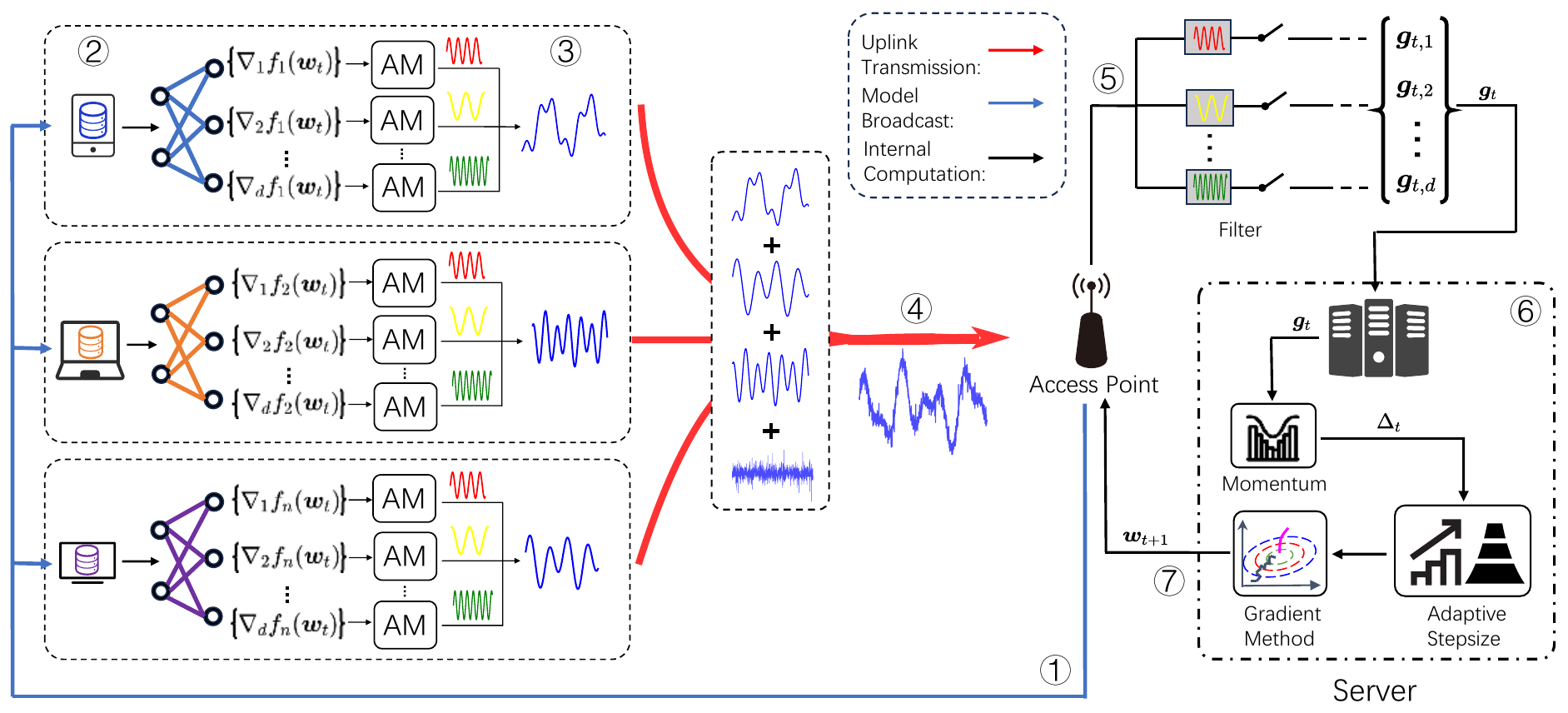}}
  \caption{An overview of the over-the-air edge learning system. The local gradients of each client are uploaded via analog transmissions, which automatically aggregate at the RF front end of the access point. The server filters out this radio signal to obtain a noisy global gradient, which is further processed and used to improve the global model. Steps of the model training in a typical communication round are numbered accordingly.}
  \label{fig:FL_WN}
\end{figure*}

\section{System Model}
We consider the federated edge learning system depicted in Fig.~1, which comprises one server and $N$ clients. 
The clients communicate with the server through wireless channels over shared spectrum. 
Every client $n \in \{1, ..., N\}$ has its local dataset $\mathcal{D}_n=\left\{\left(\boldsymbol{x}_i \in \mathbb{R}^d, y_i \in \mathbb{R}\right)\right\}_{i=1}^{m_n}$ with size $\left|\mathcal{D}_n\right|=m_n$, in which $\boldsymbol{x}_i$ and $y_i$ denote the data sample and its corresponding label, respectively. We assume the local datasets are statistically independent across the clients. 

The goal of all the entities in this system is to jointly train a statistical model using data from all the clients without exchanging their private data. 
More specifically, the edge server needs to coordinate with the clients to solve an optimization problem of the following form~\cite{MaMMooRam:16}:
\begin{align}\label{equ:obj_func}
\min_{\boldsymbol{w} \in \mathbb{R}^d} \quad f(\boldsymbol{w})=\frac{1}{N} \sum_{n=1}^N f_n(\boldsymbol{w}),
\end{align}
where $f_n(\cdot): \mathbb{R}^d \rightarrow \mathbb{R}$ is the local empirical loss function of client $n$, constructed from its own dataset $\mathcal{D}_n$, and $\boldsymbol{w} \in \mathbb{R}^d$ is the global model parameter.
We denote the optimal solution to \eqref{equ:obj_func} by $\boldsymbol{w}^*$, i.e., 
\begin{align}
    \boldsymbol{w}^*=\arg \min_{ \boldsymbol{w} \in \mathbb{R}^d } f(\boldsymbol{w}).
\end{align}

The server employs federated learning for the model training to obtain $\boldsymbol{w}^*$ while concurrently helping preserve the clients' data privacy. 
Specifically, the clients train their models locally and upload the intermediate gradients to the server. 
The server aggregates the clients' gradients and further improves the global model. 
Then, the server broadcasts the model to all the clients for another round of local training. Such interactions repeat until the global model converges. 

Due to the limited spectral resources, the efficiency of the federated training procedure is often throttled by the communication bottleneck.
For instance, under the digital communication-based model exchange paradigm, the server can only select a portion of clients for parameter uploading in each communication round \cite{yang_tcom_fl_scheduling}, which is cumbersome when the number of clients is large. 
The next section introduces a model training framework that addresses this bottleneck using the A-OTA computation method. Additionally, it is devised based on the adaptive gradient descent method, which can accelerate the model training process. Owing to these two attributes, we call our method \textit{adaptive over-the-air federated learning (ADOTA-FL)}. 

\section{Adaptive Over-the-Air Federated Learning}\label{sec:OppD2D}
This section details the design of the ADOTA-FL method. 
This method employs A-OTA computations in the global gradient aggregation stage, substantially reducing access latency and facilitating (theoretically unlimited) algorithm scalability. 
Moreover, the scheme integrates commonly used adaptive optimization techniques (e.g., AdaGrad and Adam) for learning rate optimization.
The general steps are summarized in Algorithm~\ref{Alg:Gen_FL}, and more details are provided below. 

\subsection{Gradient Aggregation Over-the-Air}
In this part, we briefly describe the approach of leveraging analog transmissions for automatic parameter aggregations over the air; a more in-depth elaboration can be found in \cite{YangChen:22JSTSP} or \cite{SerCoh:20TSP}.

Without loss of generality, we assume that model training has progressed to the $t$-th communication round, upon which the clients have just received the global model $\boldsymbol{w}_t$ from the edge server.{\footnote{Since the server can broadcast its signal at a high transmit power, we assume all the clients can successfully receive and decode the global model. }} 
Subsequently, each client $n$ updates its local gradient $\nabla f_n( \boldsymbol{w}_t )$ by taking the global model as an input. 
Then, each client performs amplitude modulation with this gradient vector (i.e., modulates its gradient vector entry-by-entry onto the magnitudes of some radio bases) using a common set of orthogonal baseband waveforms. 
More precisely, the analog signal of client $n$ can be expressed as:
\begin{align}
    x^t_n(s) = \langle \boldsymbol{\varphi}(s), \nabla f_n( \boldsymbol{w}_t ) \rangle,
\end{align}
where $\boldsymbol{\varphi}(s) = ({\varphi}_1(s), ..., {\varphi}_d(s))^\top$, $s \in [0, \tau]$, is a vector of radio waveforms, with its entries satisfying
\begin{align}
    &\int_0^{\tau} \varphi_i^2(s) ds = 1, \quad~~i = 1, 2, ..., d, \\
    &\int_0^{\tau} \varphi_i(s) \varphi_j(s) ds = 0, \qquad~ ~ i \neq j,
\end{align}
where $\tau$ represents the signal duration. 

The analog waveforms $\{ x_n^t(s) \}_{ n=1 }^N$, once constructed, are transmitted by the clients simultaneously to the edge server. 
Owing to the superposition property of electromagnetic waves, the radio signal received by the edge server has the following form: 
\begin{align}
    y(s) = \sum_{n=1}^N h_{ n, t } P_n x_n^t(s) + \xi(s),
\end{align}
where $h_{ n, t }$ is the channel fading experienced by client $n$, $P_n$ is the transmit power, set to compensate for the path loss{\footnote{Note that the path loss of each client varies slowly over time and can be accurately estimated via long-term averages of the received signal strength \cite{Li:06TWC}.}}, and $\xi(s)$ represents the electromagnetic interference. 
In this work, we assume the channel fading is independently and identically distributed (i.i.d.) across clients, with mean and variance being $\mu_{ \mathrm{c} }$ and $\sigma_{ \mathrm{c} }^2$, respectively. 
Moreover, we assume $\xi(s)$ follows a symmetric $\alpha$-stable distribution, which is a well-recognized model to characterize interference's statistical behavior in wireless networks \cite{ClaPed:21LCOMM,Mid:1977TEMC,WinPin:09JPROC}.

This received signal will be passed through a bank of match filters, with each branch tuning to $\varphi_i(s)$, $i = 1, 2, .., d$ (cf. Step 5 in Fig.~\ref{fig:FL_WN}). On the output side, the server obtains the following vector: 
\begin{equation}\label{eq:ota_agg}
\boldsymbol{g}_{t} = \frac{1}{N} \sum_{n=1}^{N} h_{ n, t } \nabla f_{n}\left( \boldsymbol{w}_{t} \right)+\boldsymbol{\xi}_{t},
\end{equation}
in which $\boldsymbol{\xi}_{t}$ denotes a $d$-dimensional random vector, with i.i.d. entries following the $\alpha$-stable distribution. 
Notably, the vector given in \eqref{eq:ota_agg} is a distorted version of the globally aggregated gradient, which will be further processed by the edge server and used for improving the global model.

{\remark{\textit{Although $\boldsymbol{g}_{t}$ is corrupted by channel fading and interference, it serves as an unbiased (but scaled) estimate of the objective function's gradient since $\mathbb{E}[ \boldsymbol{g}_{t} ] = \mu_{ \mathrm{c} } \nabla f( \boldsymbol{w}_{t} )$. However, due to the effects of the heavy-tailed interference, the variance of $\boldsymbol{g}_{t}$ is unbounded.}}}

{\remark{\textit{The underlying assumption in over-the-air gradient aggregation is that the clients are synchronized and can align their analog waveforms in time. In practice, achieving strict synchronization across a large number of clients could be challenging (or even unattainable).
In this case, one may use the method developed in \cite{ShaGun:22TWC, hells_optmis_2023} to cope with the signal misalignment issue; implementing the OTA framework in conjunction with OFDM modulation (where the Fourier basis serves as the set of orthogonal waveforms) is another solution for the synchronization issue, where the impacts of time-misalignment can be mitigated using a cyclic prefix.}}}

\begin{algorithm}[t!]
\caption{Adaptive Over-the-Air FL (ADOTA-FL)}
\textbf{Input:} Initial delay vector $\boldsymbol{v}_{-1}$, initial global model $\boldsymbol{w}_{0}$, communication round $T$, step size $\eta$
\begin{algorithmic}[1]
\FOR { $t = 0, 1, 2, ..., T-1$ }
\FOR{ each client $n \in N$ \textbf{in parallel} }
\Statex \qquad \# \textit{Train model locally and upload gradients}
\State $\nabla f_n\left(\boldsymbol{w}_t\right) \leftarrow \textsc{ClientUpdate}\left(n, \boldsymbol{w}_t\right)$
\ENDFOR
\State  $\boldsymbol{g}_t = \frac{1}{N} \sum_{n=1}^N h_{ n, t } \nabla f_n\left(\boldsymbol{w}_t\right)+\boldsymbol{\xi}_t$ 
\State  $\boldsymbol{\Delta}_t = \beta_1 \boldsymbol{\Delta}_{t-1} + \left( 1 - \beta_1 \right) \boldsymbol{g}_t$ 
\State  \colorbox{cyan!30}{$\boldsymbol{v}_t=\boldsymbol{v}_{t-1}+ \boldsymbol{\Delta}_t^\alpha$ \qquad \qquad \qquad \quad \quad\Comment{(AdaGrad)}}
\State  \colorbox{pink!30}{$\boldsymbol{v}_t=\beta_2 \boldsymbol{v}_{t-1}+\left(1-\beta_2\right) \boldsymbol{\Delta}_t^\alpha$ \, \qquad \qquad  \Comment{(Adam)}}
\State  $\boldsymbol{w}_{t+1}=\boldsymbol{w}_t-\eta  \frac{ \boldsymbol{\Delta}_t }{\sqrt[\alpha]{\boldsymbol{v}_t+\varepsilon}}$ 
\ENDFOR
\Statex \textbf{Output:} $\boldsymbol{w}_T$
\end{algorithmic} \label{Alg:Gen_FL}

\begin{algorithmic}[1]
\REQUIRE{\textsc{ClientUpdate}}($n$, $\boldsymbol{w}_t$)
\REQ{$\boldsymbol{w}_t$ broadcast to client $n$}
\STATE{Training locally using gradients method to get $\nabla f_n\left(\boldsymbol{w}_t\right)$}
\RETURN  { $\nabla f_n\left(\boldsymbol{w}_t\right)$}
\end{algorithmic}

\end{algorithm}

\subsection{Adaptive Gradient Updating}
Using $\boldsymbol{g}_{t}$, the server can update the global model at the end of the communication round \cite{YangChen:22JSTSP}. 
However, due to channel fading and interference perturbations, the globally aggregated gradient may experience significant distortion, deteriorating the performance of ordinary gradient descent-based methods. 

To alleviate the effects of such distortions, we store and update an intermediate global model as follows:
\begin{align} \label{equ:IntMdt_term}
\boldsymbol{\Delta}_t = \beta_1 \boldsymbol{\Delta}_{t-1} + \left(1-\beta_1\right) \boldsymbol{g}_t, 
\end{align}
where $0 \leq \beta_1 <1$ is a parameter controlling the relative weight of historical information and the newly acquired information. 
In essence, operation \eqref{equ:IntMdt_term} leverages a momentum-like approach to smooth out the fluctuation in the aggregated gradient. 
As training rounds continue, this update method collects the exponential moving average of the gradient and, therefore, can cope with the distortion.

Similarly to the adaptive optimization method  \cite{Adam2015,reddi2020adaptive}, we accumulate the gradient information to construct a vector $\boldsymbol{v}_t$ that automatically decays the step size in the model training.
Specifically, the AdaGrad-based method updates $\boldsymbol{v}_t$ as follows:
\begin{align}\label{eq:update_adag}
\boldsymbol{v}_t= \boldsymbol{v}_{t-1}+ \boldsymbol{\Delta}_t^\alpha
\end{align}
where $\alpha$ is the tail index in the interference distribution. 
For the Adam-like approach, such a vector is updated by
\begin{align}\label{eq:update_ema_v}
\boldsymbol{v}_t=\beta_2 \boldsymbol{v}_{t-1}+\left(1-\beta_2\right) \boldsymbol{\Delta}_t^\alpha,
\end{align}
where we slightly abuse notation by defining $\boldsymbol{\Delta}_t^\alpha = (|\boldsymbol{\Delta}_{t,1}|^\alpha, ..., |\boldsymbol{\Delta}_{t,d}|^\alpha)^\top$ and 
$0 < \beta_2 < 1$ is a hyper-parameter that controls the level of amortization the algorithm imposes on historical information (the smaller the $\beta_2$, the more historical information is taken into account).
We term the ADOTA algorithms pertaining to these two cases the AdaGrad-OTA and Adam-OTA, respectively.

Finally, using $\boldsymbol{v}_t$ and $\boldsymbol{\Delta}_t$, we update the global model as
\begin{equation}\label{eq:ota_final} 
\boldsymbol{w}_{t+1}=\boldsymbol{w}_t-\eta\frac{ \boldsymbol{\Delta}_t }{\sqrt[\alpha]{\boldsymbol{v}_t + \varepsilon}},
\end{equation}
where $\eta$ is the learning rate and $\varepsilon$ is a positive constant added to each entry of $\boldsymbol{v}_t$ to prevent ill-conditioning. 
Moreover, the division and $\alpha$-root operation in \eqref{eq:ota_final} are performed entry-wise.
As such, each element of the (re-weighted) gradient has its learning rate related to the historical information of the model training. 
This ensures that the stepsize of the different dimensions of the parameter is influenced by its value, whereas the smaller the parameter's total value, the higher the corresponding stepsize on that dimension.

The updated new global model $\boldsymbol{w}_{t+1}$ will be broadcast to all the clients for the next round of local computing. The clients and server will repeat this process for multiple rounds until the global model converges. 

{\remark{
\textit{The proposed algorithm requires estimating the interference's tail index $\alpha$, which can be efficiently accomplished via the approach in \cite{hvtail2}.}
}}

\section{Convergence Analysis }\label{sec:Analysis}
This section derives analytical expressions for the convergence rate of the ADOTA algorithms. 
Notably, existing convergence theorems for (stochastic) gradient descent under heavy-tailed noise cannot be applied directly to ADOTA because the stepsize is a random variable and depends on the historical information of the trajectory.
As a result, the technical derivations involve a number of subtle operations. 
Most proofs and mathematical derivations have been relegated to the appendix for better readability.

\subsection{Preliminaries}
To facilitate the analysis, we make the following assumptions.

\begin{assumption}
The objective function $f$ is lower bounded by a constant $f_*$, i.e.,
\begin{align}
 f(\boldsymbol{w}) \geq f_*, \qquad \forall \boldsymbol{w} \in \mathbb{R}^d.
\end{align}
\end{assumption}

\begin{assumption}
All the gradients of functions $f_n(\boldsymbol{w})$, $n \in \{1, 2, \ldots, N\}$, are bounded, namely, there exists a constant $C$ such that
\begin{align}
\|\nabla f_n(\boldsymbol{w})\|_{\infty} \leq C, \qquad \forall \boldsymbol{w} \in \mathbb{R}^d
\end{align} 
where $\Vert \boldsymbol{x} \Vert_\infty = \max_{ 1 \leq i \leq d } \{ \vert x_i \vert \}$ is the L$_\infty$ norm.
\end{assumption}

\begin{assumption}
The function $f$ is $L$-smooth under the $\alpha$-norm, i.e., for a constant $L$, the following holds 
\begin{align}
\|\nabla f(\boldsymbol{u})-\nabla f(\boldsymbol{v})\|_\alpha \leq L\|\boldsymbol{u} - \boldsymbol{v} \|_\alpha, \quad \forall \boldsymbol{u}, \boldsymbol{v} \in \mathbb{R}^d.
\end{align} 
\end{assumption}

The assumptions above have been used in various federated learning applications \cite{rmsprop,reddi2020adaptive,XiaZhu:21ICC}. 

Because each element of $\boldsymbol{\xi}_t$ has a finite $\alpha$-{\rm th} moment, we assume that the $\alpha$-{\rm th} moment of $\boldsymbol{\xi}_t$ is upper bounded by a constant $G$, namely, 
\begin{align}
\mathbb{E}\left[\left\| \boldsymbol{\xi}_t \right\|_\alpha^\alpha\right] \leq G, \qquad \forall t \in \mathbb{N}.
\end{align}

In addition, we introduce two notions that will be frequently referred to in our technical proofs. 
Specifically, we define the signed power of a vector and the complimentary of the tail index, respectively, as follows. 
\begin{definition}
For a vector $\boldsymbol{w}=\left(w_1, \ldots, w_d\right)^\top \in \mathbb{R}^d$, we define its signed power as follows:
\begin{align}
\boldsymbol{w}^{\langle\alpha\rangle}=\left(\operatorname{sgn}\left(w_1\right)\left|w_1\right|^\alpha, \ldots, \operatorname{sgn}\left(w_d\right)\left|w_d\right|^\alpha\right)^\top
\end{align}
where $\operatorname{sgn}(x) \in\{-1,+1\}$ takes the sign of the variable $x$.
\end{definition}

\begin{definition}
For the tail index $\alpha \in (1, 2]$, we define its compliment as another scalar $\gamma > 0$, satisfying 
\begin{align}
\frac{1}{\alpha} + \frac{1}{\gamma} = 1.
\end{align}    
\end{definition}

\subsection{Convergence Rate of AdaGrad Over-the-Air}

We begin by analyzing the convergence rate of OTA training model using the AdaGrad algorithm. First, we present a particular property about the smoothness of the objective function.

\begin{lemma} 
For any two vectors $\boldsymbol{u}, \boldsymbol{v} \in \mathbb{R}^d$, the objective function $f$ satisfies 
\begin{align}
f\left(\boldsymbol{u}\right)  &\leq f\left(\boldsymbol{v}\right)+\left\langle\nabla f\left(\boldsymbol{v}\right), \boldsymbol{u}-\boldsymbol{v}\right\rangle\nonumber\\
&\qquad \quad~\, +\frac{L}{2}\left(\frac{1}{\alpha} \|\boldsymbol{u}-\boldsymbol{v}\|_\alpha^\alpha + \frac{1}{\gamma}\|\boldsymbol{u}-\boldsymbol{v}\|_\gamma^\gamma\right).
\end{align}
\end{lemma}
\begin{IEEEproof}
See Appendix A.
\end{IEEEproof}

Next, we lay out three lemmas that we will use to prove the technical results. 

\begin{lemma} Given $\alpha \in[1,2]$, for any $\boldsymbol{u}, \boldsymbol{v} \in \mathbb{R}^d$, the following holds:
\begin{align}
\|\boldsymbol{u}+\boldsymbol{v}\|_\alpha^\alpha \leq\|\boldsymbol{u}\|_\alpha^\alpha + \alpha \langle\boldsymbol{u}^{\langle\alpha-1\rangle}, \boldsymbol{v} \rangle + 4\|\boldsymbol{v}\|_\alpha^\alpha.
\end{align}
\end{lemma}
\begin{IEEEproof}
Please refer to \cite{YangChen:22JSTSP}.
\end{IEEEproof}

\begin{lemma} 
Given a sequence of non-negative numbers $\{a_n\}$, we have
\begin{align}
\sum_{j=0}^n \frac{a_j}{b_j+\varepsilon} \leq  \ln \left(1+\frac{b_n}{\varepsilon}\right),
\end{align}
where $b_n=\sum_{i=0}^n  a_i$. 
\end{lemma}
\begin{IEEEproof}
See Appendix B.
\end{IEEEproof}

We now have in place all the essential building blocks out of which we can construct the convergence rate of the AdaGrad-OTA algorithm. 
This result is presented in the following.

\begin{theorem} \label{thm:AdaGrad_OTA_ConvRt}
Under the employed edge learning framework, the AdaGrad-OTA algorithm converges as 
\begin{align} \label{equ:AdaGrad-OTA_ConvRt}
& \frac{1}{T} \sum_{t=0}^{T-1} \mathbb{E}\left[\left\|\nabla f\left(\boldsymbol{w}_{t}\right)\right\|_2^2\right] \leq \frac{ \left( f\left(\boldsymbol{w}_0\right)-f_* \right) \sqrt[\alpha]{\Upsilon } }{ \eta \left( \mu_{\mathrm{c}} - 1 \right) \sqrt[\gamma]{T}} 
\nonumber\\
&+\! \frac{\left(\frac{ \eta^{ \frac{ \alpha }{ \gamma } } L }{ \alpha}+\frac{ \eta^{ \frac{ \gamma }{ \alpha } } L  }{ \gamma} + \! \Upsilon^{\frac{1}{\alpha}} \! + \! \frac{ 1 }{ \sqrt[\alpha]{\varepsilon} }  \right) \! d  \!\sqrt[\alpha]{\Upsilon }   }{ 2 ( \mu_{ \mathrm{c} } - 1 ) \sqrt[\gamma]{T} } \ln \! \left(1+\frac{ \Upsilon T }{\varepsilon}\right) 
\end{align}
where $\Upsilon$ is given by
\begin{align} \label{equ:Upsln}
\Upsilon = 4G + \frac{ d^{1 - \frac{\alpha}{2}} \left(\mu_{\mathrm{c}}^2 + \sigma_{\mathrm{c}}^2 \right)^{\frac{\alpha}{2}}  C^\alpha }{N^{ \frac{\alpha}{2}}}. 
\end{align}
\end{theorem}
\begin{IEEEproof}
See Appendix C.
\end{IEEEproof}

This result characterizes the effects of several system factors, i.e., channel fading, electromagnetic interference, model dimension, and adaptive stepsize, on the convergence rate. 
Several remarks are in order based on this theorem. 

{
\remark{
    \textit{Unlike existing results \cite{YangChen:22JSTSP,YanCheQue:22TSP,XuYanZha:22ICC} on the convergence analysis of OTA federated learning with heavy-tailed interference, which normally assume the objective function to be strongly convex and smooth, the analysis presented in Theorem~\ref{thm:AdaGrad_OTA_ConvRt} only requires smoothness of the loss function and hence is applicable to even the setting of (deep) neural networks. 
    As a result, the theoretical framework established in this paper can support a wide range of new studies in OTA machine learning. 
    }
}

\remark{
    \textit{Despite the aggregated gradient being distorted by channel fading and electromagnetic interference (whose variance is unbounded), the proposed AdaGrad-like algorithm assures that the trained model converges into a local region around the stationary points, even under a non-convex objective function. 
    }
}

\remark{
    \textit{The convergence rate is governed by $\mathcal{O}( \frac{ \ln{T} }{ T^{ 1/\gamma } } ) = \mathcal{O}( \frac{ \ln{T} }{ T^{ 1 - \frac{1}{\alpha} } } )$, indicating that the tail index $\alpha$ plays a decisive role in the convergence performance. More concretely, the smaller the $\alpha$, namely, the heavier the interference's tails, the slower the algorithm converges. 
    }
}

\remark{
    \textit{The tail index also profoundly affects the multiplicative terms in the convergence rate, whereas a decrease in $\alpha$ increases the multipliers, slowing down the training convergence. }
}

\remark{
    \textit{When the interference follows a Gaussian distribution, i.e., $\alpha=2$, the AdaGrad-OTA algorithm converges on the order of $\mathcal{O}( \frac{ \ln{T} }{ \sqrt{T} } )$, which retrieves the convergence rate of standard AdaGrad in a federated learning system with digital communication-based parameter exchanges \cite{AdaConvBound, SimpConv}. }
}

\remark{
\textit{The analysis in \eqref{equ:AdaGrad-OTA_ConvRt} also illuminates the training efficiency in the noiseless model upload scenario, if we abuse the constraint of the tail index by having $\alpha \rightarrow \infty$ and $\sigma_{ \mathrm{c} } = 0$, where the corresponding convergence rate is $\mathcal{O}( \frac{\ln T}{ T } )$. }
}

\remark{
\textit{
The model size, i.e., dimension $d$, has a marked influence on the convergence performance. Since expanding the model size directly slows down the convergence rate, while this effect is on the multiplicative terms.
}
}

\remark{
    \textit{In the presence of large variations in channel fading, i.e., an increase in $\sigma_{ \mathrm{c} }$, the communications would frequently encounter deep fades, which inflict additional fluctuations in the training process and slow down the convergence (this is quantitatively reflected by the increase in $\Upsilon$). }
}

\remark{
\textit{With more clients participating in the system, i.e., increasing $N$, the impact of channel fading on the gradient aggregation can be alleviated through the averaging operation (since $\Upsilon$ decreases). 
Therefore, scaling up the system benefits the model training. This observation aligns with the previous discoveries in OTA federated learning \cite{SerCoh:20TSP,YangChen:22JSTSP} 
}
}

}
\begin{figure*}[htbp]
    \centering
     
    \begin{subfigure}[t]{0.3\textwidth}
          \centering
          \includegraphics[width=6.1cm]{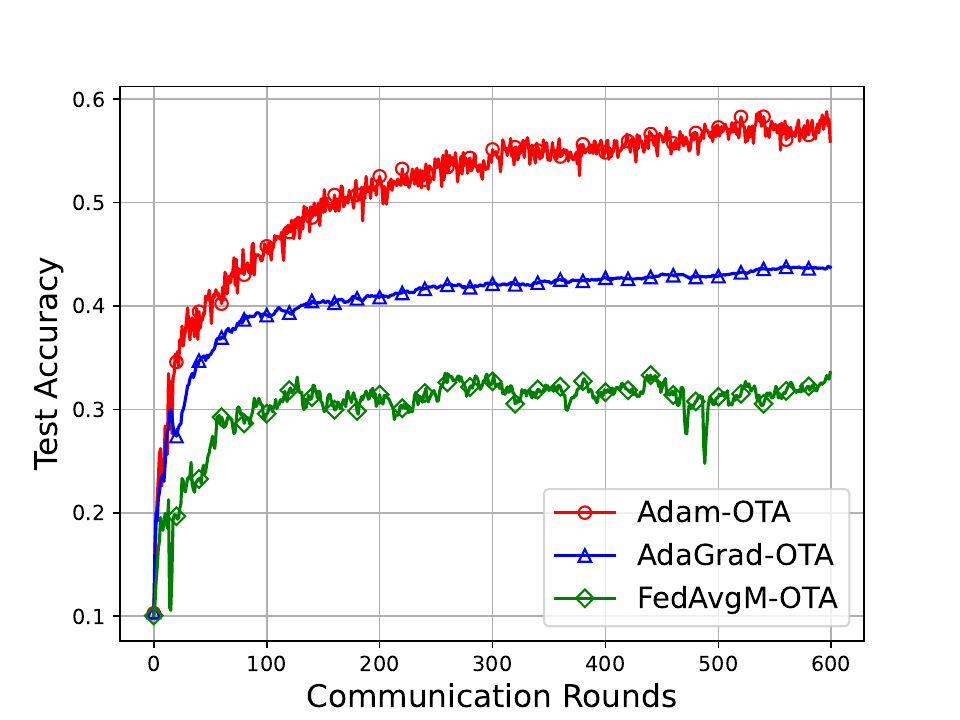}
            \centering
            \subcaption[1]{}
            \vspace{-0.5cm}
            \label{fig:res-18acc}
    \end{subfigure}
    \begin{subfigure}[t]{0.3\textwidth}
            \centering
            \includegraphics[width=6.1cm]{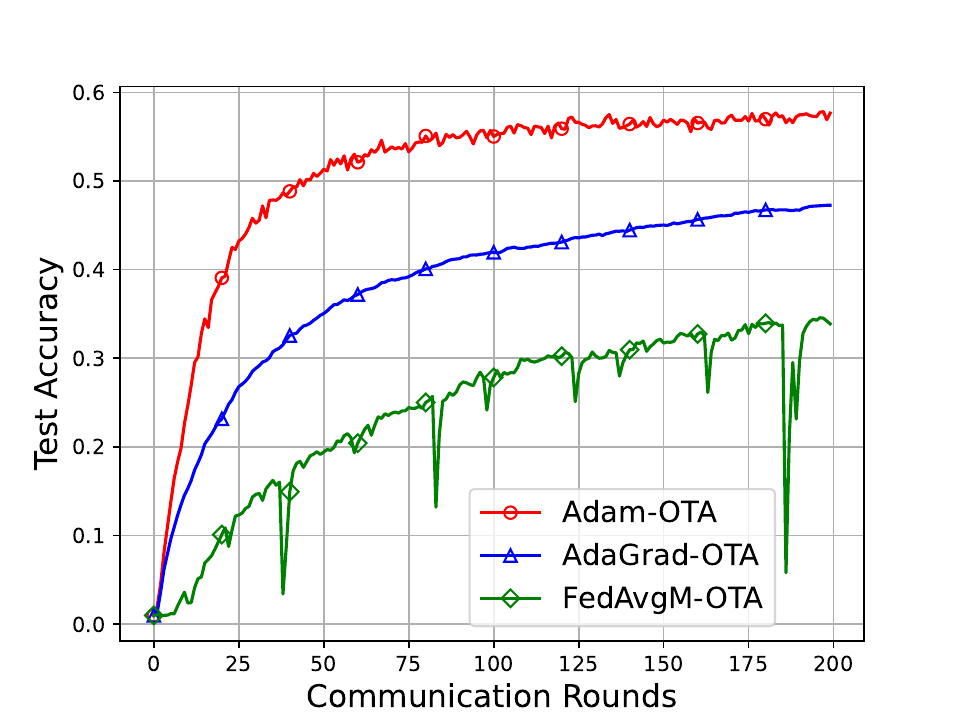}
            \vspace{-0.5cm}
            \subcaption[2]{}
            \label{fig:res-34acc}
    \end{subfigure}
    \begin{subfigure}[t]{0.3\textwidth}
            \centering
            \includegraphics[width=6.1cm]{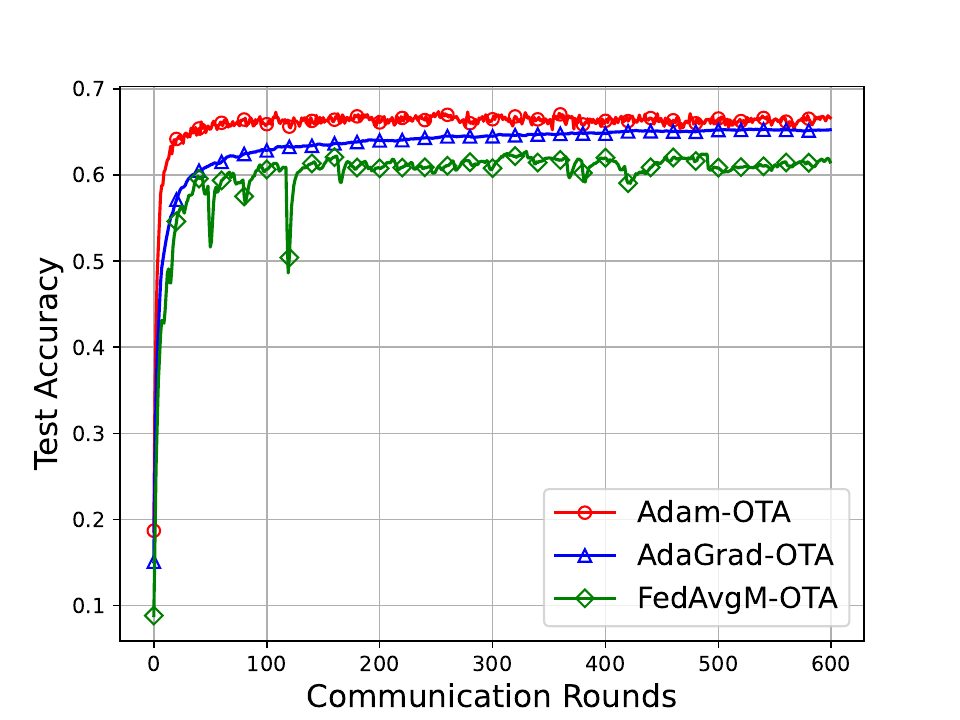}
            \vspace{-0.5cm}
            \subcaption[3]{}
            \label{fig:res-lgacc}
    \end{subfigure}

    \begin{subfigure}[t]{0.3\textwidth}
            \centering
            \includegraphics[width=6.1cm]{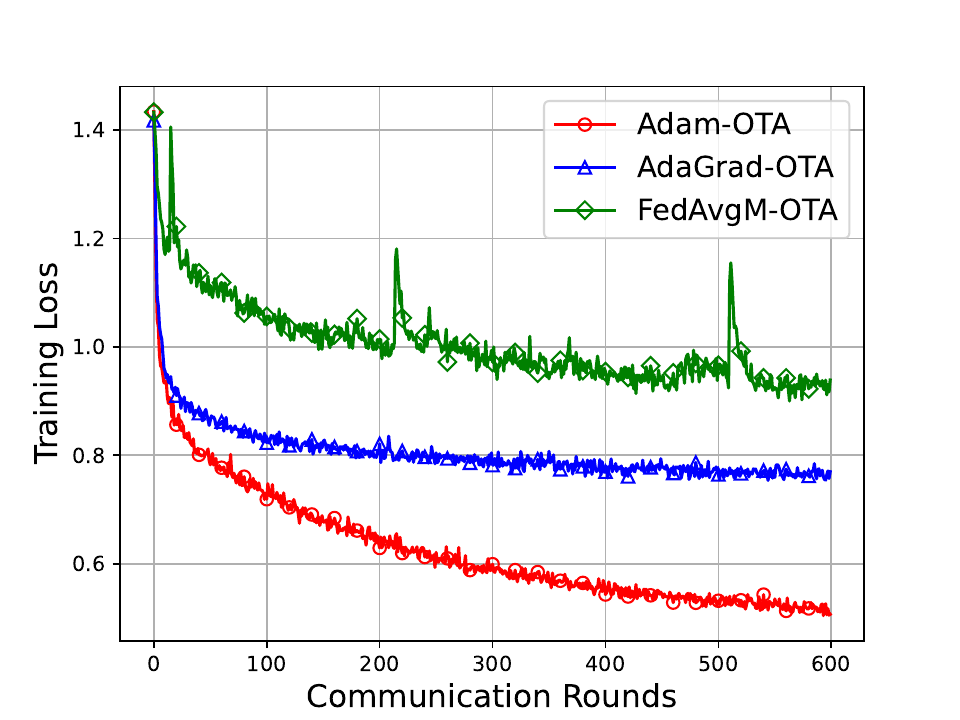}
            \vspace{-0.5cm}
            \subcaption[4]{}
            \label{fig:res-18ls}
    \end{subfigure}
    \begin{subfigure}[t]{0.3\textwidth}
          \centering
          \includegraphics[width=6.1cm]{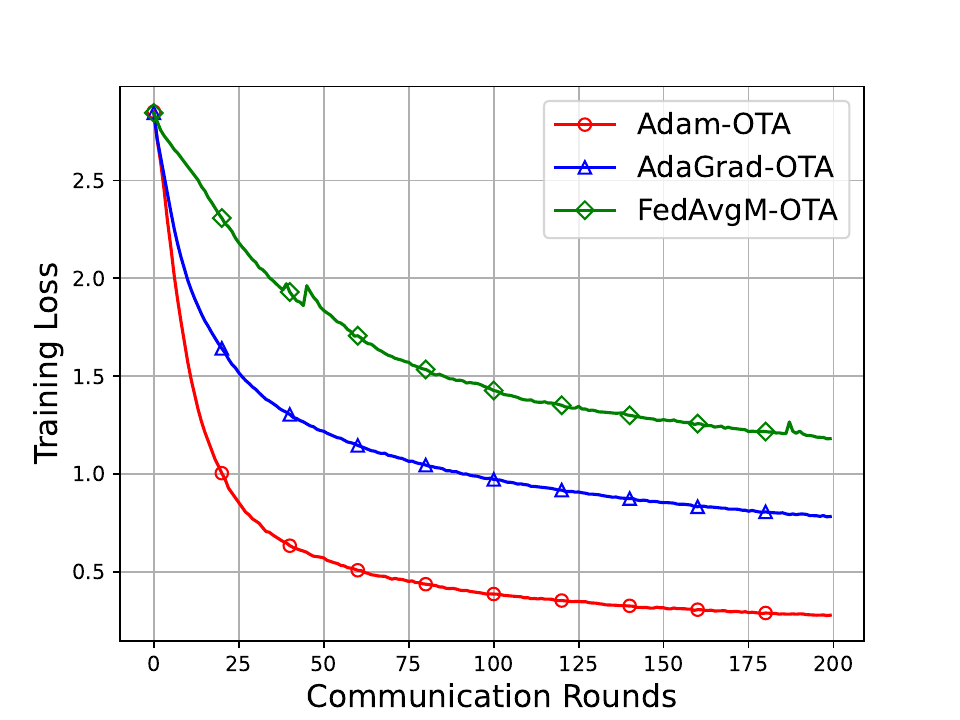}
            \vspace{-0.5cm}
        \subcaption[5]{}
        \label{fig:res-34ls}
    \end{subfigure}
    \begin{subfigure}[t]{0.3\textwidth}
            \centering
            \includegraphics[width=6.1cm]{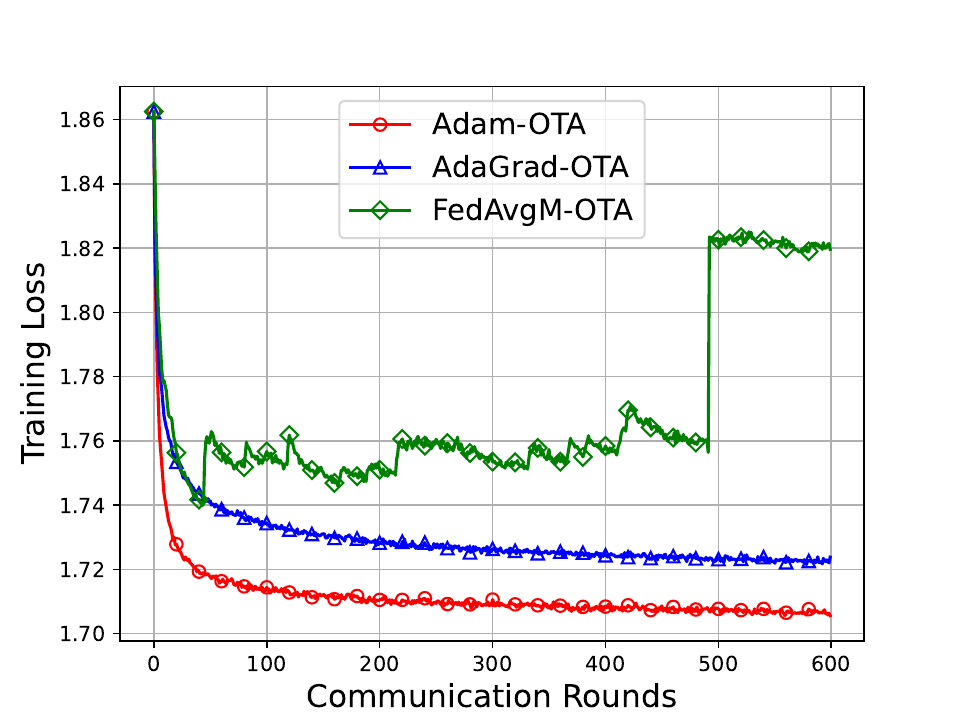}
            \vspace{-0.5cm}
            \subcaption[6]{}
            \label{fig:res-lgls}
    \end{subfigure}
    \caption{Performance comparison of the test accuracy and training loss of different tasks with non-i.i.d. data partition \textit{Dir}=0.1 under heavy tail index $\alpha = 1.5$. Here (a) and (d) are for ResNet-18 on the CIFAR-10 dataset, (b) and (e) are for ResNet-34 on the CIFAR-100 dataset, and (c) and (f) are for logistic regression on the EMNIST dataset.}
    \label{fig:res-total}
\end{figure*}

\subsection{Convergence Rate of Adam Over-the-Air}
In this subsection, we extend the analytical framework developed above to derive the convergence rate of OTA model training under an Adam-like algorithm. To begin with, we introduce the following lemma, serving as a stepping stone for the subsequent analysis. 

\begin{lemma} 
Given a sequence of non-negative numbers $\{a_n\}$ and a constant $0 < \phi \leq 1$, we have
\begin{align}
\sum_{j=0}^n \frac{a_j^2}{b_j+\varepsilon} \leq \frac{1}{1-\phi} \ln \left(1 + \frac{b_n}{\varepsilon}\right) -\frac{n \ln \phi}{1-\phi},
\end{align}
where $b_n = (1 - \phi) \sum_{i=0}^n \phi^{n-i} a_i$.
\end{lemma}

\begin{IEEEproof}
See Appendix D.
\end{IEEEproof}

\begin{theorem} \label{thm:Adam_OTA_ConvRt}
Under the employed edge learning framework, the Adam-OTA algorithm converges as
\begin{align} \label{equ:Adam-OTA_ConvRt}
&\frac{1}{T} \sum_{t=0}^{T-1} \mathbb{E}\left[\left\|\nabla f\left(\boldsymbol{w}_{t}\right)\right\|_2^2\right] \leq 
\frac{ \left( f\left(\boldsymbol{w}_0\right)-f_* \right) \sqrt[\alpha]{\Upsilon } }{ \left( \mu_{\mathrm{c}} + \beta_2 - 1 \right) \eta  {T}} 
\nonumber\\
&+ \frac{ \left(\frac{ \eta^{ \frac{ \alpha }{ \gamma } } L }{ \alpha  } + \frac{\left(1-\beta_2\right)^{\frac{1}{\gamma}}\eta^\gamma L + \gamma\left(1-\beta_2\right)^{\gamma-2}\Upsilon^{\frac{1}{\alpha}}}{ \gamma \left(1-\beta_2\right)^{\gamma+\frac{1}{\gamma}-2}} + \frac{ \left(1-\beta_2\right)}{ \sqrt[\alpha]{ \varepsilon } } \right) d \sqrt[\alpha]{\Upsilon} }{ 2 ( \mu_{ \mathrm{c} } + \beta_2 - 1 ) } 
\nonumber\\
& \times  \left( \frac{1}{T} \ln \left(1 + \frac{ \Upsilon }{\varepsilon}\right) - \ln \beta_2 \right) 
\end{align}
where $\Upsilon$ is given in \eqref{equ:Upsln}.
\end{theorem}
\begin{IEEEproof}
See Appendix E.
\end{IEEEproof}

Similar observations as in the previous subsection can also be obtained from \eqref{equ:Adam-OTA_ConvRt}. 
In addition, this result reveals two distinct properties of the Adam-OTA algorithm.

{ \remark{
\textit{The Adam-like OTA model training attains a convergence rate on the order of $\mathcal{O}( \frac{ 1 }{ T } )$, which brings a substantial improvement compared to AdaGrad-OTA (cf. Theorem~1). 
This gain can be attributed primarily to the exponential moving average, which makes the adaptive level more dependent on the recent 
 gradient than the previous ones. }
}
}

{\remark{
\textit{The control factor $\beta_2$ affects the convergence rate in a non-linear manner. Hence, it can be adjusted to accelerate the convergence of the algorithm. }
}
}

\section{ Simulation Results }\label{sec:NumResult}
In this section, we conduct experiments to examine the efficacy of our proposed A-DOTA framework across various system configurations.
We start by detailing the setup of our experiments.
Then, we assess the effectiveness of our proposed ADOTA-FL schemes by comparing them to a state-of-the-art baseline. 
Finally, we investigate the impact of different system parameters on the algorithms' performance, including the number of participating clients, the tail index of the interference distribution, and data heterogeneity (note that although our analysis was conducted under the assumption of i.i.d. training dataset, we will also evaluate the algorithms on non-i.i.d. dataset to explore how they perform in that situation).

\subsection{Setup}
We evaluate the performance of our proposed ADOTA-FL framework on two data sets, CIFAR-10 and CIFAR-100~\cite{cifar10ref}, using model architectures ResNet-18 and ResNet-34~\cite{resnet}, respectively. 
We also assess its performance in training a logistic regression model on the EMNIST~\cite{emnistref} dataset. 
To demonstrate the efficacy of our proposed method, we adopted state-of-the-art FedAvgM as the baseline under the same system setup for performance comparison\footnote{We did not include the performance of FedAvg because FedAvgM outperforms it in most real-world data settings.}. 

For data partition, we use the widely adopted symmetric Dirichlet distribution for heterogeneous local data simulation, in which the degree of data heterogeneity across local clients is controlled by the concentration parameter \textit{Dir}.  
Unless otherwise stated, we use concentration parameter \textit{Dir} = 0.1 for non-i.i.d. data partition.
For experiments on CIFAR-10 (resp. CIFAR-100), we use $N=100$ (resp. $N=50$). For experiments on EMNIST, we use $N=50$.
Regarding the channel models in the A-OTA system, we employ the Rayleigh fading to model the channel gain and use symmetric $\alpha$-stable distribution to characterize the interference, where we assign the average channel gain and tail index as $\mu_{\mathrm{c}} = 1$ and $\alpha = 1.5$, respectively. If not specified otherwise, the scale of the interference is $0.1$.
The experiments are implemented with Pytorch on NVIDIA RTX 3090 GPU.
Regarding the performance evaluation, we evaluate the generalization and convergence performance of our proposed A-DOTA framework by using the test accuracy of the global model and averaged training loss across clients as evaluation metrics, respectively.

\subsection{Performance evaluation} 
In Fig.~\ref{fig:res-total}, we compare the test accuracy and training loss of our proposed method with the baseline over different datasets under the A-OTA FL system configurations.
From Figs.~\ref{fig:res-18acc} and \ref{fig:res-18ls}, which illustrate the training results based on non-i.i.d. CIFAR-10 dataset,
we observe a remarkable performance gain by adopting the A-DOTA scheme.
Specifically, compared to the FadAvgM-OTA baseline, the AdaGrad-OTA approach significantly speeds up the training convergence. 
Additionally, it increases the test accuracy by more than $10\%$.
This gain becomes more pronounced by adopting the Adam-OTA algorithm, where the convergence rate is further enhanced, and the test accuracy attains an almost two-fold improvement. 
Moreover, the convergence curve (of both the test accuracy and training loss) of FedAvgM-OTA exhibits notable fluctuations, especially with the occurrence of impulsive interference.
In contrast, the AdaGrad-OTA and Adam-OTA methods are able to mitigate the effect of channel disturbances by adapting the stepsizes.
Similar observations are also evident from Figs.~\ref{fig:res-34acc} and \ref{fig:res-34ls}, in which the model training is performed on the CIFAR-100 dataset, validating the advantage of our proposed method in tasks with larger model sizes and higher computational complexities.  

In addition to performance evaluation with deep neural networks (i.e., ResNet-18/34)-based tasks, 
we also conducted experiments with convex objective functions by carrying out logistic regression tasks on the EMNIST dataset.
As shown in Fig.~\ref{fig:res-lgacc} and \ref{fig:res-lgls}, our proposed algorithm outperforms the state-of-the-art baseline, which presents the same conclusion as given in the corresponding results of CIFAR-10 and CIFAR-100. 
Notably, the results are also consistent with the findings in the theoretical analysis.
It is also noteworthy that in training tasks with convex objectives, the convergence and robustness performance of the FedAvgM is significantly affected by the noisy gradient aggregation in A-OTA FL with heavy-tailed noise.

\begin{figure}
  \centering
  \begin{subfigure}[b]{0.95\linewidth}
    \includegraphics[width=\linewidth]{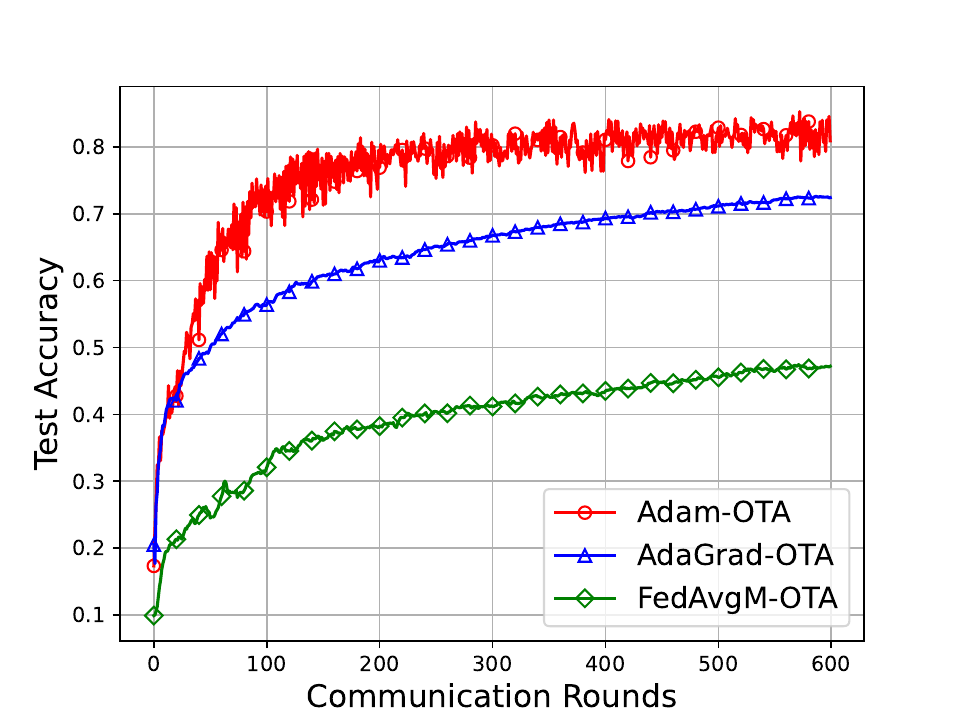} 
    \subcaption[5]{}
    \label{fig:18acc-nols}
  \end{subfigure}
  
  \begin{subfigure}[b]{0.95\linewidth}
    \includegraphics[width=\linewidth]{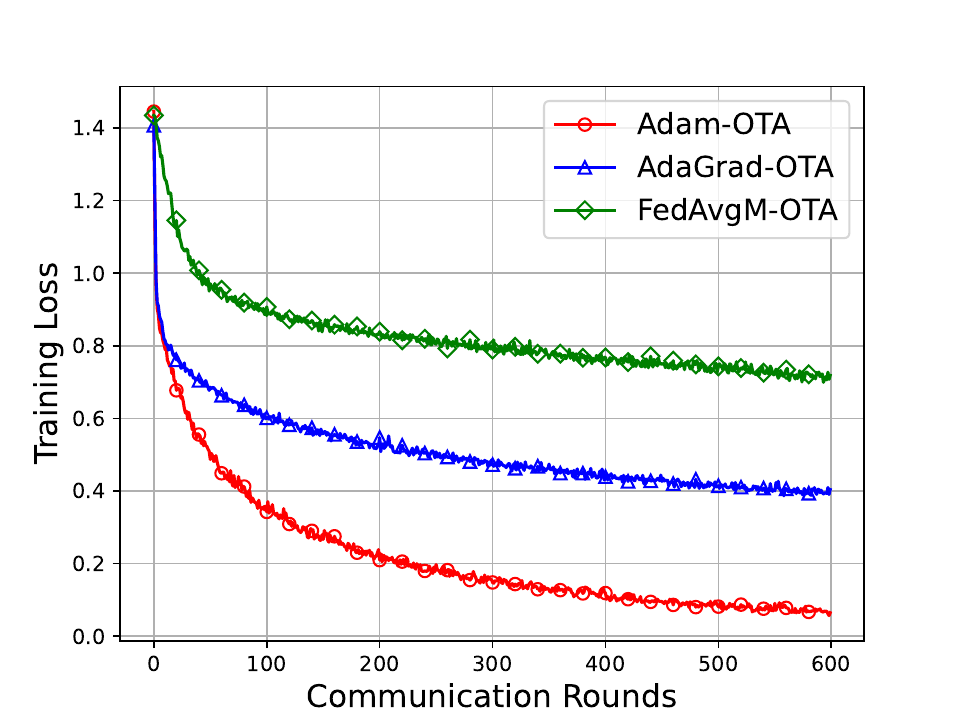}
    \subcaption[5]{}
    \label{fig:18loss-nols}
  \end{subfigure}
  
  \caption{Performance comparison for test accuracy and training loss under tail index $\alpha = 1.8$ and scale = $0.01$, of training a ResNet-18 on the CIFAR-10 dataset.}
  \label{fig:18-nols}
  \vspace{-8pt}
\end{figure}

We also investigated the performance with different channel noise A-OTA computing setups in Fig.~\ref{fig:18-nols}, in which the tail index of heavy-tailed noise is set to $1.8$ and the scale of the noise is set to $0.01$.
Fig.~\ref{fig:18acc-nols} and \ref{fig:18loss-nols} depict the test accuracy and training loss evaluation performance over the CIFAR-10 dataset with identical FL system configurations in Fig.~\ref{fig:res-18ls}, respectively. 
The results reveal that our proposed methods consistently outperform the baseline with various A-OTA setups, illustrating the generalized superiority. 

In summary, our proposed ADOTA-FL scheme outperforms the state-of-the-art OTA baseline with respect to both generalizability and convergence across diverse tasks and heterogeneous data setups. 
We attribute such consistent outperformance to the adaptive optimization in our proposed frameworks, which could alleviate the performance degradation due to noisy gradient transmission and aggregation, brought by the channel fading and interference in A-OTA FL.

\subsection{Effects of hyper-parameters} 

\begin{figure}[t!]
  \centering{\includegraphics[width=0.95\columnwidth]{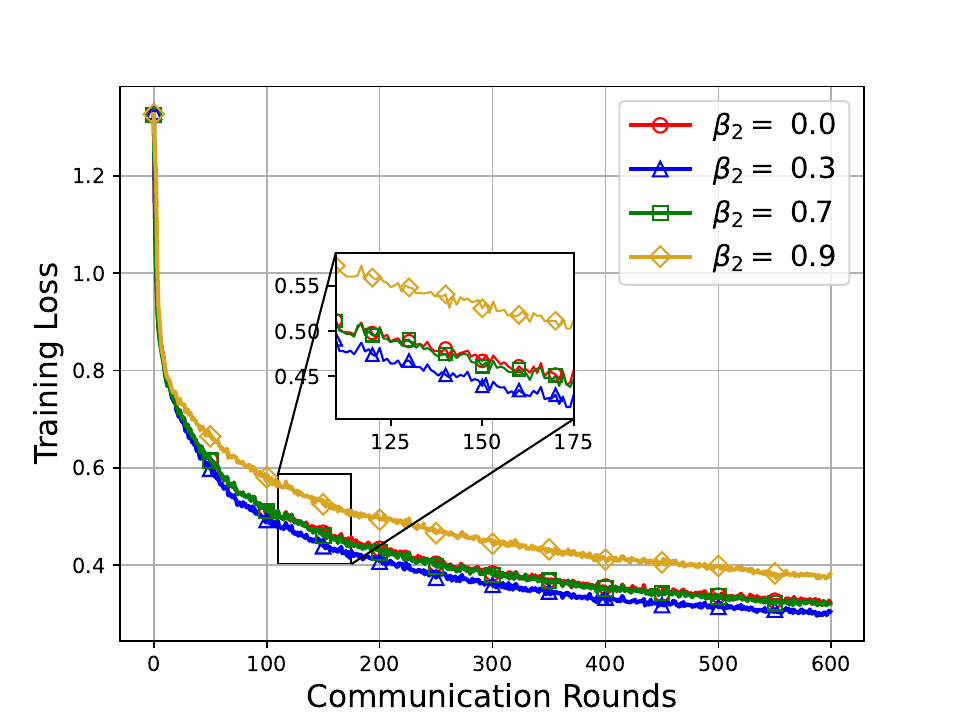}}
  \caption{Performance comparison for training loss with $\beta_1 = 0$ and non-i.i.d. data partition \textit{Dir} = 0.1 under different $\beta_2$. We use the Adam-OTA method to train ResNet-18 on CIFAR-10.}  
  \label{fig:res-beta}
\end{figure}

In Fig. \ref{fig:res-beta}, we explore the impact of the control factor $\beta_2$ on the Adam-OTA algorithm by varying its value, where the convergence curves are plotted under different values of $\beta_2$. 
Firstly, we observe that a well-chosen $\beta_2$, i.e., $\beta_2=0.3$, can effectively enhance the convergence rate. 
The findings are consistent with Remark~14, i.e., appropriately adjusting $\beta_2$ is instrumental in expediting training convergence.
Secondly, we find that setting $\beta_2$ to extreme regimes (either too large or too small) may slow down the overall training process (e.g., $\beta_2=0.9$) due to misusing historical gradient information.

\subsection{Effects of system parameters} 
In this subsection, we focus on the generalization capability of our proposed ADOTA-FL framework on different system setups. For conciseness, all the results are obtained from the evaluations on the AdaGrad-OTA method. Specifically, we focus on the impacts of different tail indices of the heavy-tailed channel interference, different system scales, and different data heterogeneity settings.

\begin{figure}[t!]
  \centering{\includegraphics[width=0.95\columnwidth]{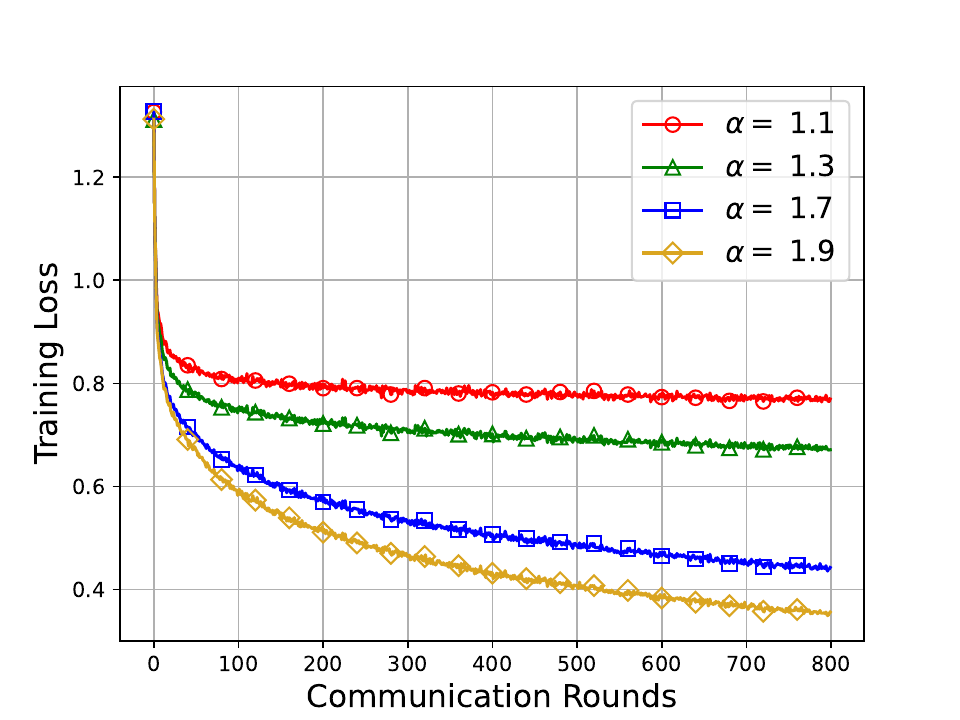}} 
  \caption{Performance comparison for training loss with non-i.i.d. data partition \textit{Dir} = $0.1$ under different $\alpha$. We use the AdaGrad-OTA method to train ResNet-18 on CIFAR-10.}  
  \label{fig:res-tail}
  \vspace{-8pt}
\end{figure}

\textbf{Performance with different tail indices:} Fig.~\ref{fig:res-tail} plots the training loss of ADOTA-FL as a function of the communication round under different values of tail index $\alpha$.
It demonstrates that as the tail index increases, the training performance of ADOTA-FL improves with a faster convergence rate. 
The numerical results are aligned with Remark~6, confirming that channel noise with a smaller tail index signifies a slower decay rate. 
Since our algorithm incorporates adaptive descent to mitigate the impact of extreme values, the overall performance of the gradient descent is inevitably influenced. Consequently, a diminishing tail index (lower $\alpha$) corresponds to a deceleration in the convergence rate.

\begin{figure}[t!]
  \centering{\includegraphics[width=0.95\columnwidth]{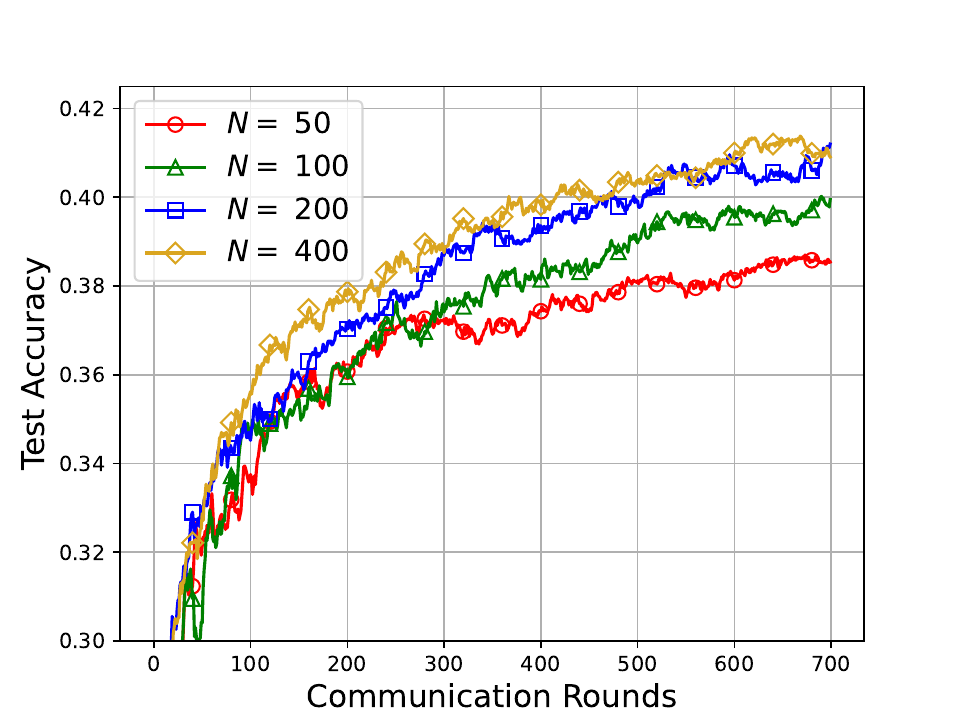}}
  \caption{Performance comparison for test accuracy with non-i.i.d. data partition \textit{Dir} = $0.2$ under different total numbers of clients $N$. We use the AdaGrad-OTA method to train ResNet-18 on CIFAR-10.}  
  \label{fig:res-num}
  \vspace{-8pt}
\end{figure}

\textbf{Performance with different system scales:} 
We further examine the performance of ADOTA-FL under different system scales (i.e., the total number of clients $N$ in the A-OTA FL system), as illustrated in Fig.~\ref{fig:res-num}.
This figure illustrates a distinctive phenomenon inherent to A-OTA FL systems, wherein an increased number of participating clients correlates positively with the system's performance enhancement. This phenomenon stems from the utilization of A-OTA computing, in which all clients can concurrently transmit their locally computed gradients during each communication round. In contrast to the conventional digital communication FL that necessitates scheduling, this approach substantially augments the volume of information aggregated in each round, consequently amplifying the generalization performance. Furthermore, the aggregation of imperfect gradients from an increased number of clients mitigates the deleterious effects of channel interference and fading.

\begin{figure}[t!]
  \centering{\includegraphics[width=0.95\columnwidth]{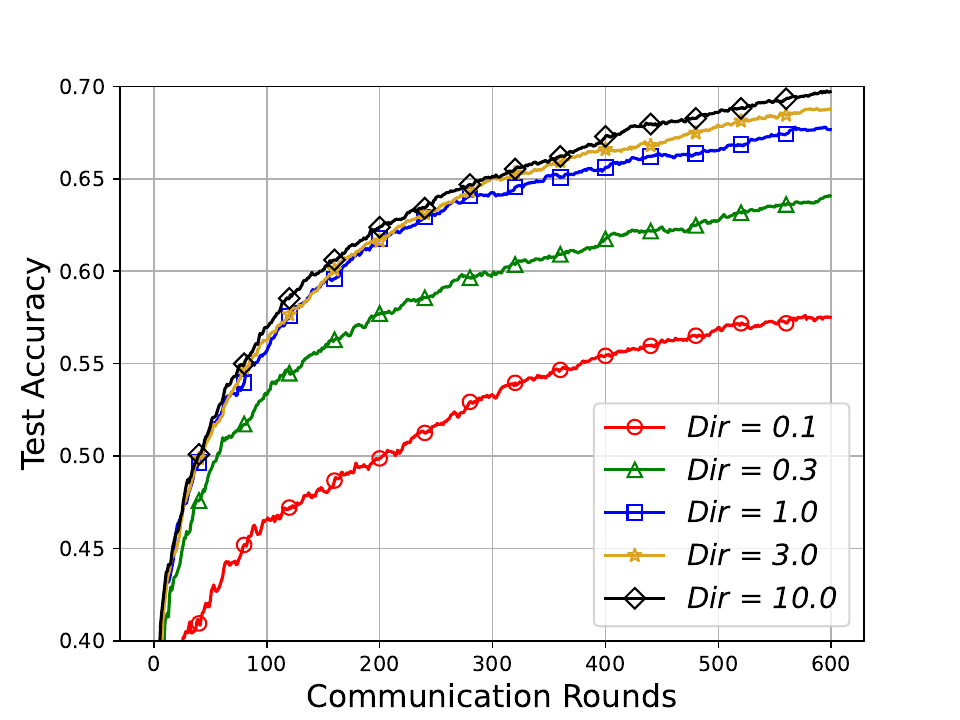}}
  \caption{Performance comparison for test accuracy under different \textit{Dir}. We use the AdaGrad-OTA method to train ResNet-18 on CIFAR-10.}  
  \label{fig:res-alpha}
  \vspace{-8pt}
\end{figure}

\textbf{Performance with different degrees of data heterogeneity:} We also investigate the performance of the ADOTA-FL with different degrees of data heterogeneity. In our experiments, we use different \textit{Dir} to control the degrees of heterogeneity in the data, in which a smaller value of \textit{Dir} indicates a more non-i.i.d. data partition. 
Fig.~\ref{fig:res-alpha} shows the impact of \textit{Dir} on the training performance, which indicates that the convergence performance will slow down as the degree of data heterogeneity increases (i.e., a smaller value of \textit{Dir}). The AdaGrad-like method presents stable performance with diverse heterogeneous data settings.

\section{ Conclusion }\label{sec:Conclusion}
We have proposed the ADOTA-FL framework, which incorporates adaptive gradient methods into the OTA-FL system, enhancing the robustness of the model training process.
We have developed the A-OTA FL versions of AdaGrad and Adam algorithms, which accumulate historical gradient information to update the stepsize in each global iteration, combating the detriments of channel fading and interference in analog transmissions.
We have derived the convergence rates of the proposed ADOTA-FL methods for general non-convex loss functions, accounting for the effects of key system factors such as the number of participating clients, channel fading, and electromagnetic interference. 
Our analysis reveals that the level of heavy-tailedness of the interference distribution plays a dominant role in the convergence rate of AdaGrad-based schemes, where the heavier the tail, the slower the algorithm converges. 
In contrast, the Adam-OTA method is more resilient to channel distortions and converges faster. 
We have conducted extensive experiments to examine the efficacy of our proposed methods.
The results show that the ADOTA FL methods outperform a state-of-the-art baseline across various system configurations, corroborating the proposed approaches' effectiveness and validating our theoretical findings.

Although this paper focuses on the AdaGrad and Adam-like methods, the developed framework can be extended to study the integration of other adaptive optimizers with the OTA FL system. 
In this context, developing an OTA FL model update strategy that tackles the heavy-tailed effects arising from data heterogeneity \cite{XiaCheLiu:23NuerIPS} and electromagnetic interference is a particularly interesting extension. 
Another concrete direction for future study of this technique is investigating its effectiveness in dealing with mobility issues \cite{FenYanHu:21TWC} in a multi-cell OTA FL network.

\begin{appendix}

\subsection{Proof of Lemma 1}

Following Assumption 3, we have
\begin{align}
&|f(\boldsymbol{u})-f(\boldsymbol{v})-\langle\nabla f(\boldsymbol{v}), \boldsymbol{u}-\boldsymbol{v}\rangle| \nonumber\\
& \leq \int_0^1|\langle\nabla f(\boldsymbol{v}+s(\boldsymbol{u}-\boldsymbol{v}))-\nabla f(\boldsymbol{v}), \boldsymbol{u}-\boldsymbol{v}\rangle| d s \nonumber\\
& \stackrel{(a)}{\leq} \int_0^1\|\nabla f(\boldsymbol{v}+s(\boldsymbol{u}-\boldsymbol{v}))-\nabla f(\boldsymbol{v})\|_\alpha \|\boldsymbol{u}-\boldsymbol{v}\|_\gamma d s \nonumber\\
& \leq \int_0^1 s L\|\boldsymbol{u}-\boldsymbol{v}\|_\alpha \|\boldsymbol{u}-\boldsymbol{v}\|_\gamma d s \nonumber\\
& \stackrel{(b)}{\leq} \frac{L}{2}\left(\frac{1}{\alpha} \|\boldsymbol{u}-\boldsymbol{v}\|_\alpha^\alpha + \frac{1}{\gamma}\|\boldsymbol{u}-\boldsymbol{v}\|_\gamma^\gamma\right)
\end{align}
where (a) follows from Hölder's inequality and (b) follows from Young's inequality. The proof is completed by expanding the absolute value operation and moving to the right-hand side.

\subsection{Proof of Lemma 3}
Given $b_n>a_n \geq 0$, for all $n \in \mathbb{N}^*$, we have 
\begin{align}
\frac{a_j}{b_j+\varepsilon}  \leq \ln \left(b_j+\varepsilon\right)-\ln \left(\varepsilon+ b_{j-1}\right), \quad \forall j\in \mathbb{N}^*.
\end{align}
The above results in a telescoping series. 
The proof is completed by summing over all $j \in \{ 0, 1, \cdots, n \}$.

\subsection{Proof of Theorem 1}
Using Lemma 1 (with $\gamma \geq \alpha$) and taking the conditional expectation (where the condition is on all historical information up to iteration $t$) of $f\left(\boldsymbol{w}_{t+1}\right)$, we have

\begin{align}
\mathbb{E}_t\left[f\left(\boldsymbol{w}_{t+1}\right)\right] &\leq  f\left(\boldsymbol{w}_t\right)-\eta\mathbb{E}_{t}\left[\frac{\nabla_i f\left(\boldsymbol{w}_{t}\right) \boldsymbol{g}_{t, i}}{\sqrt[\alpha]{\boldsymbol{v}_{t, i}+\varepsilon}}\right] \nonumber\\
&+\left(\frac{\eta^\alpha L}{2 \alpha}+\frac{\eta^\gamma L}{2 \gamma}\right) \sum_{i=1}^d \mathbb{E}_t\left[\frac{|\boldsymbol{g}_{t, i}|^\alpha}{ \boldsymbol{v}_{t, i} + \varepsilon }\right].
\label{eq:t1}
\end{align}

For ease of exposition, we denote an auxiliary variable by:
\begin{align}
\boldsymbol{\tilde{v}}_{t, i}= \boldsymbol{v}_{t-1, i}+\mathbb{E}_{t}\left[|\boldsymbol{g}_{t, i}|^\alpha\right].
\end{align}
Then, we have the following:
\begin{align} \label{equ:AdaG_drv_intmdt_inrprdct}
&\mathbb{E}_{t}\left[\frac{\nabla_i f\left(\boldsymbol{w}_{t}\right) \boldsymbol{g}_{t, i}}{\sqrt[\alpha]{\boldsymbol{v}_{t, i}+\varepsilon}}\right]=\mathbb{E}_{t}\left[\frac{\nabla_i f\left(\boldsymbol{w}_{t}\right) \boldsymbol{g}_{t, i}}{\sqrt[\alpha]{\boldsymbol{\tilde{v}}_{t, i}+\varepsilon}}\right] \nonumber\\
&+\mathbb{E}_{t}\Bigg[\underbrace{\nabla_i f\left(\boldsymbol{w}_{t}\right) \boldsymbol{g}_{t, i}\left(\frac{1}{\sqrt[\alpha]{\boldsymbol{v}_{t, i}+\varepsilon}}-\frac{1}{\sqrt[\alpha]{\boldsymbol{\tilde{v}}_{t, i}+\varepsilon}}\right)}_S \Bigg].
\end{align}

Note that $\mathbb{E}_{t}\left[\boldsymbol{g }_{t, i}\right] = \mu_{\mathrm{c}} \nabla_i f\left(\boldsymbol{w}_{t}\right)$, and  $\boldsymbol{g}_{t, i}$ and $\boldsymbol{\boldsymbol{\tilde{v}}_{t, i}}$ are independent of each other, we can calculate the first term on the right-hand side of \eqref{equ:AdaG_drv_intmdt_inrprdct} as 
\begin{align}
\mathbb{E}_{t}\left[\frac{\nabla_i f\left(\boldsymbol{w}_{t}\right) \boldsymbol{g}_{t, i}}{\sqrt[\alpha]{\boldsymbol{\tilde{v}}_{t, i}+\varepsilon}}\right]=\frac{ \mu_{\mathrm{c}} \nabla_i f\left(\boldsymbol{w}_{t}\right)^2}{\sqrt[\alpha]{\boldsymbol{\tilde{v}}_{t, i}+\varepsilon}}.
\end{align}

Next, we need to bound $\mathbb{E}_t[S]$ in \eqref{equ:AdaG_drv_intmdt_inrprdct}. 
We begin by expanding the expression of $S$, which gives an initial upper bound as follows:
\begin{align}
S \leq   \frac{\nabla_i f\left(\boldsymbol{w}_{t}\right) \boldsymbol{g}_{t, i} \left(\mathbb{E}_{t}\left[|\boldsymbol{g}_{t, i}|^\alpha\right]-|\boldsymbol{g}_{t, i}|^\alpha\right)}{ \sqrt[\alpha]{ \left(\boldsymbol{v}_{t, i} \!+\! \varepsilon\right) \left(\boldsymbol{\tilde{v}}_{t, i} \!+\! \varepsilon\right) } \big( \left(\boldsymbol{v}_{t, i} \!+\! \varepsilon\right)^{1/\gamma} \!+\! \left(\boldsymbol{\tilde{v}}_{t, i} \!+\! \varepsilon\right)^{1/\gamma} \big) }.
\end{align}

Then, leveraging the following relationships: 
\begin{align}
&\left(\boldsymbol{v}_{t, i}+\varepsilon\right)^{1/\gamma}+\left(\boldsymbol{\tilde{v}}_{t, i}+\varepsilon\right)^{1/\gamma} 
\nonumber\\
&\qquad \qquad \qquad \geq \max \big( \left(\boldsymbol{v}_{t, i}+\varepsilon\right)^{1/\gamma}, \left(\boldsymbol{\tilde{v}}_{t, i}+\varepsilon\right)^{1/\gamma} \big)
\end{align}
and 
\begin{align}
\left|\mathbb{E}_{t}\left[|\boldsymbol{g}_{t, i}|^\alpha\right]-|\boldsymbol{g}_{t, i}|^\alpha\right| \leq     \mathbb{E}_{t}\left[|\boldsymbol{g}_{t, i}|^\alpha\right]+|\boldsymbol{g}_{t, i}|^\alpha,
\end{align}
we have 
\begin{align}
|S| & \leq \underbrace{|\nabla_i f\left(\boldsymbol{w}_{t}\right) \boldsymbol{g}_{t, i}| \frac{\mathbb{E}_{t}\left[|\boldsymbol{g}_{t, i}|^\alpha\right]}{\left(\boldsymbol{v}_{t, i}+\varepsilon\right)^{1/\alpha}(\boldsymbol{\tilde{v}}_{t, i}+\varepsilon)}}_{S_1} \nonumber\\
&+\underbrace{|\nabla_i f\left(\boldsymbol{w}_{t}\right) \boldsymbol{g}_{t, i}| \frac{|\boldsymbol{g}_{t, i}|^\alpha}{(\boldsymbol{v}_{t, i}+\varepsilon) \left(\boldsymbol{\tilde{v}}_{t, i}+\varepsilon\right)^{1/\alpha}}}_{S_2}. \label{adag:stotal}
\end{align}

Subsequently, we bound ${S_1}$ and ${S_2}$, respectively. 
First of all, we adopt the following inequality: 
\begin{align}\label{lm:totalsqr}
x y \leq \frac{\lambda}{2} x^2+\frac{y^2}{2 \lambda}, \qquad \forall \lambda>0, x, y \in \mathbb{R}, 
\end{align}
and apply it to $S_1$, with $\lambda$, $x$, and $y$ respectively setting to
\begin{align}
&\lambda=\left(\boldsymbol{\tilde{v}}_{t, i}+\varepsilon\right)^{1/\alpha}, \qquad
x=\frac{|\nabla_i f\left(\boldsymbol{w}_{t}\right)|}{\left(\boldsymbol{\tilde{v}}_{t, i}+\varepsilon\right)^{1/\alpha}}, \nonumber\\
& y=\frac{|\boldsymbol{g}_{t, i}| \mathbb{E}_{t}\left[|\boldsymbol{g}_{t, i}|^\alpha\right]}{\left(\boldsymbol{\tilde{v}}_{t, i}+\varepsilon\right)^{1/\gamma} \left(\boldsymbol{v}_{t, i}+\varepsilon\right)^{1/\alpha}}.
\end{align}
Taking a conditional expectation yields
\begin{align}
\mathbb{E}_{t}[{S_1}] \leq \frac{\nabla_i f\left(\boldsymbol{w}_{t}\right)^2}{2 \left(\boldsymbol{\tilde{v}}_{t, i}+\varepsilon\right)^{1/\alpha}}+\frac{\mathbb{E}_{t}\left[|\boldsymbol{g}_{t, i}|^\alpha\right]^{\frac{1}{\alpha}}}{2} \mathbb{E}_{t}\left[\frac{|\boldsymbol{g}_{t, i}|^\alpha}{\boldsymbol{v}_{t, i}+\varepsilon}\right].
\end{align}

The next step is to bound $\mathbb{E}_{t}\left[|\boldsymbol{g}_{t, i}|^\alpha\right]$. By applying Lemma~3 and recognizing the fact that $|\boldsymbol{g}_{t, i}|^\alpha \leq \|\boldsymbol{g}_{t}\|_\alpha^\alpha$, we have
\begin{align}
\mathbb{E}_{t}\left[\|\boldsymbol{g}_{t}\|_\alpha^\alpha\right] & \leq \mathbb{E}_{t}\left[ \Big \|\frac{1}{N} \sum_{n=1}^N h_{t, n} \nabla f_n\left(\boldsymbol{w}_{t}\right) \Big\|_\alpha^\alpha\right]+4\mathbb{E}_{t}\left[\|\boldsymbol{\xi}_{t}\|_\alpha^\alpha\right]
\nonumber\\
& \stackrel{(a)}{\leq} \frac{ d^{1 - \frac{\alpha}{2}}  }{N^{\alpha}} \sum_{n=1}^N  \mathbb{E}_{t}\left[\left( \| h_{t, n} \nabla f_n\left(\boldsymbol{w}_{t}\right)\|_2^2\right)^{\frac{\alpha}{2}}\right] + 4G  \nonumber\\
& \stackrel{(b)}{\leq} \frac{ d^{1 - \frac{\alpha}{2}}  }{N^{\alpha}} \sum_{n=1}^N  \mathbb{E}_{t}\left[ \| h_{t, n} \nabla f_n\left(\boldsymbol{w}_{t}\right)\|_2^2\right]^{\frac{\alpha}{2}} + 4G   \nonumber\\
& \leq d^{1 - \frac{\alpha}{2}} \left(\mu_{\mathrm{c}}^2 + \sigma_{\mathrm{c}}^2 \right)^{\frac{\alpha}{2}} N^{-\frac{\alpha}{2}} C^\alpha + 4G,
\end{align}
where (a) and (b) follow from Hölder's inequality and Jensen’s inequality, respectively.

As such, we have 
\begin{align}
\mathbb{E}_{t}[{S_1}] \leq \frac{\nabla_i f\left(\boldsymbol{w}_{t}\right)^2}{2 \left(\boldsymbol{\tilde{v}}_{t, i}+\varepsilon\right)^{1/\alpha}}+\frac{\Upsilon^{\frac{1}{\alpha}}}{2} \mathbb{E}_{t}\left[\frac{|\boldsymbol{g}_{t, i}|^\alpha}{\boldsymbol{v}_{t, i}+\varepsilon}\right] \label{adag:s1}
\end{align}
where $\Upsilon$ is given in \eqref{equ:Upsln} and 
\begin{align}
\mathbb{E}_{t}\left[|\boldsymbol{g}_{t, i}|^\alpha\right]\leq  4G + \frac{ d^{1 - \frac{\alpha}{2}} \left(\mu_{\mathrm{c}}^2 + \sigma_{\mathrm{c}}^2 \right)^{\frac{\alpha}{2}}  C^\alpha }{N^{ \frac{\alpha}{2}}}.
\end{align}

In order to bound ${S_2}$, we can recursively apply \eqref{lm:totalsqr} with
\begin{align}
\lambda=\frac{\left(\boldsymbol{\tilde{v}}_{t, i}+\varepsilon\right)^{1/\alpha}}{ \mathbb{E}_{t}\left[|\boldsymbol{g}_{t, i}|^\alpha\right]^{\gamma-1}}, x=\frac{|\nabla_i f\left(\boldsymbol{w}_{t}\right)| |\boldsymbol{g}_{t, i}|^{\frac{\alpha}{2}}}{\sqrt[\alpha]{\boldsymbol{\tilde{v}}_{t, i}+\varepsilon}}, y=\frac{|\boldsymbol{g}_{t, i}|^{\frac{\alpha}{2}}}{\boldsymbol{v}_{t, i}+\varepsilon}
\end{align}
which yields 
\begin{align}
\mathbb{E}_{t}[{S_2}] \leq \frac{\nabla_i f\left(\boldsymbol{w}_{t}\right)^2}{2 \left(\boldsymbol{\tilde{v}}_{t, i}+\varepsilon\right)^{1/\alpha}} +\frac{1}{2 \varepsilon^{1/\alpha}} \mathbb{E}_{t}\left[\frac{|\boldsymbol{g}_{t, i}|^\alpha}{\boldsymbol{v}_{t, i}+\varepsilon}\right]. \label{adag:s2}
\end{align}

Using the fact that $\mathbb{E}_t \left[\boldsymbol{g}_{t, i}^\alpha  \right] \geq \mathbb{E}_t \left[\boldsymbol{g}_{t, i}  \right]^\alpha$ and $\boldsymbol{\tilde{v}}_{t, i} \geq 0$, and substituting \eqref{adag:s1} and \eqref{adag:s1} into \eqref{adag:stotal}, we arrive at the following: 
\begin{align}
\mathbb{E}_{t}[|S|] \leq \frac{\nabla_i f\left(\boldsymbol{w}_{t}\right)^2}{ \left(\boldsymbol{\tilde{v}}_{t, i}+\varepsilon\right)^{1/\alpha}} +\frac{\Upsilon^{\frac{1}{\alpha}} + \varepsilon^{-\frac{1}{\alpha}}}{2} \mathbb{E}_{t}\left[\frac{|\boldsymbol{g}_{t, i}|^\alpha}{\boldsymbol{v}_{t, i}+\varepsilon}\right],
\end{align}
here $\mathbb{E}_{t}[|S|] \geq |\mathbb{E}_{t}[S]|$. Expanding this absolute value inequality and substitute it into the original formula; we get 
\begin{align}
\mathbb{E}_{t}\left[\frac{\nabla_i f\left(\boldsymbol{w}_{t}\right) \boldsymbol{g}_{t, i}}{\left(\boldsymbol{v}_{t, i}+\varepsilon\right)^{1/\alpha}}\right] 
&\geq \frac{\left(\mu_{\mathrm{c}} - 1\right) \nabla_i f\left(\boldsymbol{w}_{t}\right)^2 }{ \left(\boldsymbol{\tilde{v}}_{t, i}+\varepsilon\right)^{1/\alpha}} \nonumber\\
& - \frac{\Upsilon^{\frac{1}{\alpha}} + \varepsilon^{-\frac{1}{\alpha}}}{2} \mathbb{E}_{t}\left[\frac{|\boldsymbol{g}_{t, i}|^\alpha}{\boldsymbol{v}_{t, i}+\varepsilon}\right]. 
\end{align}

Applying the above inequality into \eqref{eq:t1} and notice that $ \left(\boldsymbol{\tilde{v}}_{t, i}+\varepsilon\right)^{1/\alpha} \leq  \sqrt[\alpha]{\Upsilon \left(t+1\right)}$, we have 
\begin{align}
&\mathbb{E}_{t}\left[f\left(\boldsymbol{w}_{t+1}\right)\right] \leq f\left(\boldsymbol{w}_{t}\right)-\frac{\left(\mu_{\mathrm{c}} - 1\right)}{\sqrt[\alpha]{\Upsilon \left(t+1\right)}}
\left\|\nabla f\left(\boldsymbol{w}_{t}\right)\right\|_2^2\nonumber\\
&\qquad \qquad \qquad +\left(\frac{\eta^\alpha L}{2 \alpha}+\frac{\eta^\gamma L}{2 \gamma}\right) \sum_{i=1}^d \mathbb{E}_t\left[\frac{|\boldsymbol{g}_{t, i}|^\alpha}{\boldsymbol{v}_{t, i}+\varepsilon}\right] \nonumber\\
&\qquad \qquad \qquad \quad~ + \eta \frac{\Upsilon^{\frac{1}{\alpha}} + \varepsilon^{-\frac{1}{\alpha}}}{2} \sum_{i=1}^d\mathbb{E}_{t}\left[\frac{|\boldsymbol{g}_{t, i}|^\alpha}{\boldsymbol{v}_{t, i}+\varepsilon}\right].
\end{align}

To this end, by summing the previous inequality through $t \in \{0, 1, \ldots, T-1\}$ and taking the complete expectation, we have
\begin{align}
\mathbb{E}\left[f\left(\boldsymbol{w}_{T}\right)\right] &\leq f\left(\boldsymbol{w}_{0}\right)-\frac{\eta \left(\mu_{\mathrm{c}} - 1\right) }{ \sqrt[\alpha]{\Upsilon T}}
\sum_{t=0}^{T-1} \mathbb{E}\left[\left\|\nabla f\left(\boldsymbol{w}_{t}\right)\right\|_2^2\right] \nonumber\\
&\qquad +\left(\frac{\eta^\alpha L}{2 \alpha}+\frac{\eta^\gamma L}{2 \gamma}\right) \sum_{t=0}^{T-1} \sum_{i=1}^d \mathbb{E} \left[\frac{|\boldsymbol{g}_{t, i}|^\alpha}{\boldsymbol{v}_{t, i}+\varepsilon}\right] \nonumber\\
&\qquad\quad~ + \eta \frac{\Upsilon^{\frac{1}{\alpha}} + \varepsilon^{-\frac{1}{\alpha}}}{2} \sum_{t=0}^{T-1} \sum_{i=1}^d \mathbb{E} \left[\frac{|\boldsymbol{g}_{t, i}|^\alpha}{\boldsymbol{v}_{t, i}+\varepsilon}\right]. 
\end{align}
The proof is completed by invoking Lemma 3 and simplifying the above formula.

\subsection{Proof of Lemma 4} 
Given $b_n>\left(1 - \phi\right)a_n \geq 0$ for all $n \in \mathbb{N}^*$, we have for 
\begin{align}
\left(1-\phi\right) \frac{a_j}{b_j+\varepsilon} 
&=\ln \left(\frac{b_j+\varepsilon}{b_{j-1}+\varepsilon}\right)+\ln \left(\frac{b_{j-1}+\varepsilon}{\phi b_{j-1}+\varepsilon}\right) \nonumber\\
& \leq \ln \left(\frac{b_j+\varepsilon}{b_{j-1}+\varepsilon}\right)-\ln \phi, \quad \forall j\in \mathbb{N}^*.
\end{align}
The above inequality constitutes a telescoping series. The proof is completed by summing over all $j \in\{0, 1, \ldots,n\}$.

\subsection{Proof of Theorem 2} 

Following similar lines in the proof of Theorem 1, we can expand and bound $\mathbb{E}_t\left[f\left(\boldsymbol{w}_{t+1}\right)\right]$ in the following way:
\begin{align}
&\mathbb{E}_t\left[f\left(\boldsymbol{w}_{t+1}\right)\right] \leq f\left(\boldsymbol{w}_t\right)-\eta\mathbb{E}_{t}\left[\frac{\nabla_i f\left(\boldsymbol{w}_{t}\right) \boldsymbol{g}_{t, i}}{\sqrt[\alpha]{\boldsymbol{v}_{t, i}+\varepsilon}}\right] \nonumber\\
&+\left(\frac{\eta^\alpha L}{2 \alpha}+\frac{\eta^\gamma L}{2 \gamma \left(1-\beta_2\right)^{\gamma/\alpha-1}}\right) \sum_{i=1}^d \mathbb{E}_t\left[\frac{|\boldsymbol{g}_{t, i}|^\alpha}{\left(\boldsymbol{v}_{t, i}+\varepsilon\right)}\right]. \label{eq:t2}
\end{align}

Similarly, we separate $\mathbb{E}_{t}\left[\frac{\nabla_i f\left(\boldsymbol{w}_{t}\right) \boldsymbol{g}_{t, i}}{\sqrt[\alpha]{\boldsymbol{v}_{t, i}+\varepsilon}}\right]$ into two parts:
\begin{align}
&\mathbb{E}_{t}\left[\frac{\nabla_i f\left(\boldsymbol{w}_{t}\right) \boldsymbol{g}_{t, i}}{\sqrt[\alpha]{\boldsymbol{v}_{t, i}+\varepsilon}}\right]=\mathbb{E}_{t}\left[\frac{\nabla_i f\left(\boldsymbol{w}_{t}\right) \boldsymbol{g}_{t, i}}{\sqrt[\alpha]{\boldsymbol{\tilde{v}}_{t, i}+\varepsilon}}\right] \nonumber\\
&+\mathbb{E}_{t} \Bigg[\underbrace{\nabla_i f\left(\boldsymbol{w}_{t}\right) \boldsymbol{g}_{t, i}\left(\frac{1}{\sqrt[\alpha]{\boldsymbol{v}_{t, i}+\varepsilon}}-\frac{1}{\sqrt[\alpha]{\boldsymbol{\tilde{v}}_{t, i}+\varepsilon}}\right)}_S \Bigg].\label{adam:mid}
\end{align}

By the same operation as in Appendix~C, we get
\begin{align}
|S| & \leq \left(1-\beta_2\right)\underbrace{|\nabla_i f\left(\boldsymbol{w}_{t}\right) \boldsymbol{g}_{t, i}| \frac{\mathbb{E}_{t}\left[|\boldsymbol{g}_{t, i}|^\alpha\right]}{\left(\boldsymbol{v}_{t, i}+\varepsilon\right)^{1/\alpha}(\boldsymbol{\tilde{v}}_{t, i}+\varepsilon)}}_{S_1} \nonumber\\
&+\left(1-\beta_2\right)\underbrace{|\nabla_i f\left(\boldsymbol{w}_{t}\right) \boldsymbol{g}_{t, i}| \frac{|\boldsymbol{g}_{t, i}|^\alpha}{(\boldsymbol{v}_{t, i}+\varepsilon) \left(\boldsymbol{\tilde{v}}_{t, i}+\varepsilon\right)^{1/\alpha}}}_{S_2} . \label{adam:stotal}
\end{align}

We can bound $S_1$ and $S_2$ respectively by applying \eqref{lm:totalsqr} to ${S_1}$ with
\begin{align}
&\lambda=\left(\boldsymbol{\tilde{v}}_{t, i}+\varepsilon\right)^{1/\alpha}, x=\frac{|\nabla_i f\left(\boldsymbol{w}_{t}\right)|}{\left(\boldsymbol{\tilde{v}}_{t, i}+\varepsilon\right)^{1/\alpha}}, \nonumber\\
&y=\frac{|\boldsymbol{g}_{t, i}| \mathbb{E}_{t}\left[|\boldsymbol{g}_{t, i}|^\alpha\right]}{\left(\boldsymbol{\tilde{v}}_{t, i}+\varepsilon\right)^{1/\gamma} \left(\boldsymbol{v}_{t, i}+\varepsilon\right)^{1/\alpha}}
\end{align}
and ${S_2}$ with
\begin{align}
\lambda=\frac{\left(\boldsymbol{\tilde{v}}_{t, i}+\varepsilon\right)^{1/\alpha}}{ \mathbb{E}_{t}\left[|\boldsymbol{g}_{t, i}|^\alpha\right]^{\gamma-1}}, x=\frac{|\nabla_i f\left(\boldsymbol{w}_{t}\right)| |\boldsymbol{g}_{t, i}|^{\frac{\alpha}{2}}}{\sqrt[\alpha]{\boldsymbol{\tilde{v}}_{t, i}+\varepsilon}}, y=\frac{|\boldsymbol{g}_{t, i}|^{\frac{\alpha}{2}}}{\boldsymbol{v}_{t, i}+\varepsilon}.
\end{align}
Consequently, we have
\begin{align}
\mathbb{E}_{t}[{S_1}]\leq \frac{\nabla_i f\left(\boldsymbol{w}_{t}\right)^2}{2 \left(\boldsymbol{\tilde{v}}_{t, i}+\varepsilon\right)^{1/\alpha}}+\frac{\Upsilon^{\frac{1}{\alpha}}}{2\left(1-\beta_2\right)^{1+\frac{1}{\gamma}}}\mathbb{E}_{t}\left[\frac{|\boldsymbol{g}_{t, i}|^\alpha}{\boldsymbol{v}_{t, i}+\varepsilon}\right] \label{adam:s1}
\end{align}
and 
\begin{align}
\mathbb{E}_{t}[{S_2}]
\leq \frac{\nabla_i f\left(\boldsymbol{w}_{t}\right)^\alpha}{\alpha \left(\boldsymbol{\tilde{v}}_{t, i}+\varepsilon\right)^{1/\alpha}} +\frac{1}{2 \varepsilon^{\frac{1}{\alpha}}} \mathbb{E}_{t}\left[\frac{|\boldsymbol{g}_{t, i}|^\alpha}{\boldsymbol{v}_{t, i}+\varepsilon}\right]. \label{adam:s2}
\end{align}

Then, substituting \eqref{adam:s1} and \eqref{adam:s2} to \eqref{adam:stotal} results in
\begin{align}
\mathbb{E}_{t}\left[ |S| \right] &\leq \frac{ \left(1 - \beta_2\right) \nabla_i f\left(\boldsymbol{w}_{t}\right)^2}{ \left(\boldsymbol{\tilde{v}}_{t, i}+\varepsilon\right)^{1/\alpha}}\nonumber\\
&+\frac{\Upsilon^{\frac{1}{\alpha}} + \left(1-\beta_2\right)^{1+\frac{1}{\gamma}}\varepsilon^{-\frac{1}{\alpha}}}{2\left(1-\beta_2\right)^{\frac{1}{\gamma}}}\mathbb{E}_{t}\left[\frac{|\boldsymbol{g}_{t, i}|^\alpha}{\boldsymbol{v}_{t, i}+\varepsilon}\right].
\end{align}

Because $\mathbb{E}_{t}[|S|] \geq |\mathbb{E}_{t}[S]|$, by putting the above inequality to \eqref{adam:mid}, we get
\begin{align}
&\mathbb{E}_{t}\left[\frac{\nabla_i f\left(\boldsymbol{w}_{t}\right) \boldsymbol{g}_{t, i}}{\left(\boldsymbol{v}_{t, i}+\varepsilon\right)^{1/\alpha}}\right]  \geq \frac{\left(\mu_{\mathrm{c}} - 1 + \beta_2\right) \nabla_i f\left(\boldsymbol{w}_{t}\right)^2 }{ \left(\boldsymbol{\tilde{v}}_{t, i}+\varepsilon\right)^{1/\alpha}} \nonumber\\
&-\frac{\Upsilon^{\frac{1}{\alpha}} + \left(1-\beta_2\right)^{1+\frac{1}{\gamma}}\varepsilon^{-\frac{1}{\alpha}}}{2\left(1-\beta_2\right)^{\frac{1}{\gamma}}}\mathbb{E}_{t}\left[\frac{|\boldsymbol{g}_{t, i}|^\alpha}{\boldsymbol{v}_{t, i}+\varepsilon}\right]. 
\end{align}

Applying the above inequality into the \eqref{eq:t2}, along with the fact that $(\boldsymbol{\tilde{v}}_{t, i} + \varepsilon )^{1/\alpha} \leq  \sqrt[\alpha]{\Upsilon (1-\beta_2^{t+1} )} \leq \sqrt[\alpha]{\Upsilon}$, we have
\begin{align}
&\mathbb{E}_{t}\left[f\left(\boldsymbol{w}_{t+1}\right)\right] \leq f\left(\boldsymbol{w}_{t}\right)-\frac{\eta\left(\mu_{\mathrm{c}} - 1 + \beta_2\right)}{\sqrt[\alpha]{\Upsilon}}
\left\|\nabla f\left(\boldsymbol{w}_{t}\right)\right\|_2^2 \nonumber\\
&+\left(\frac{\eta^\alpha L}{2 \alpha}+\frac{\eta^\gamma L}{2 \gamma \left(1-\beta_2\right)^{\gamma/\alpha-1}}\right) \sum_{i=1}^d \mathbb{E}_t\left[\frac{|\boldsymbol{g}_{t, i}|^\alpha}{\boldsymbol{v}_{t, i}+\varepsilon}\right] \nonumber\\
&+ \eta \frac{\Upsilon^{\frac{1}{\alpha}} + \left(1-\beta_2\right)^{1+\frac{1}{\gamma}}\varepsilon^{-\frac{1}{\alpha}}}{2\left(1-\beta_2\right)^{\frac{1}{\gamma}}}\sum_{i=1}^d\mathbb{E}_{t}\left[\frac{|\boldsymbol{g}_{t, i}|^\alpha}{\boldsymbol{v}_{t, i}+\varepsilon}\right].
\end{align}

Summing this inequality for all $t \in\{0, 1, \ldots, T-1\}$, and taking the complete expectation yields
\begin{align}
&\mathbb{E}\left[f\left(\boldsymbol{w}_{T}\right)\right] 
\leq f\left(\boldsymbol{w}_{0}\right)-\frac{\eta \left(\mu_{\mathrm{c}} - 1 + \beta_2\right)}{\sqrt[\alpha]{\Upsilon}}
\sum_{t=0}^{T-1} \mathbb{E}\left[\left\|\nabla f\left(\boldsymbol{w}_{t}\right)\right\|_2^2\right]  \nonumber\\
&+\left(\frac{\eta^\alpha L}{2 \alpha}+\frac{\eta^\gamma L}{2 \gamma \left(1-\beta_2\right)^{\gamma/\alpha-1}}\right) \sum_{t=0}^{T-1} \sum_{i=1}^d \mathbb{E} \left[\frac{|\boldsymbol{g}_{t, i}|^\alpha}{\boldsymbol{v}_{t, i}+\varepsilon}\right] \nonumber\\
&+ \eta \frac{\Upsilon^{\frac{1}{\alpha}} + \left(1-\beta_2\right)^{1+\frac{1}{\gamma}}\varepsilon^{-\frac{1}{\alpha}}}{2\left(1-\beta_2\right)^{\frac{1}{\gamma}}} \sum_{t=0}^{T-1} \sum_{i=1}^d \mathbb{E} \left[\frac{|\boldsymbol{g}_{t, i}|^\alpha}{\boldsymbol{v}_{t, i}+\varepsilon}\right]. \label{adam-final}
\end{align}

Invoking Lemma 4 and rearranging \eqref{adam-final} gives the result.

\end{appendix}

\bibliographystyle{IEEEtran}
\bibliography{adaptive_methods_ota}

\end{document}

%% file: aconym.tex
\acrodef{CCDF}{complementary cumulative distribution function}
\acrodef{CF}{characteristic function}
\acrodef{PPP}{Poisson point processe}
\acrodef{RV}{random variable}
\acrodef{i.i.d.}{independent and identically distributed}
\acrodef{PDF}{probability distribution function}
\acrodef{CDF}{cumulative distribution function}
\acrodef{ch.f.}{characteristic function}
\acrodef{AWGN}{additive white Gaussian noise}
\acrodef{SNR}{signal-to-noise ratio}
\acrodef{LRT}{likelihood ratio test}
\acrodef{DRT}{distance ratio test}
\acrodef{GLRT}{generalized likelihood ratio test}
\acrodef{CRLB}{Cram\'{e}r-Rao lower bound}
\acrodef{CRB}{Cram\'{e}r-Rao bound}
\acrodef{ZZLB}{Ziv-Zakai lower bound}
\acrodef{ZZB}{Ziv-Zakai bound}
\acrodef{LOS}{line-of-sight}
\acrodef{ToF}{time-of-flight}
\acrodef{NLOS}{non-line-of-sight}
\acrodef{GDOP}{geometric dilution of precision}
\acrodef{GPS}{Global Positioning System}
\acrodef{FIM}{Fisher information matrix}
\acrodef{PEB}{position error bound}
\acrodef{SPEB}{squared position error bound}
\acrodef{TOA}{time-of-arrival}
\acrodef{TOF}{time-of-flight}
\acrodef{WSN}{wireless sensor network}
\acrodef{MAC}{medium access control}
\acrodef{RSS}{received signal strength}
\acrodef{WAF}{wall attenuation factor}
\acrodef{TDOA}{time difference-of-arrival}
\acrodef{RF}{radiofrequency}
\acrodef{RTT}{round-trip time}
\acrodef{AOA}{angle-of-arrival}
\acrodef{MF}{matched filter}
\acrodef{ED}{energy detector}
\acrodef{ML}{maximum likelihood}
\acrodef{MSE}{mean-square error}
\acrodef{RMSE}{root-mean-square error}
\acrodef{LEO}{localization error outage}
\acrodef{ppm}{part-per-million}
\acrodef{ACK}{acknowledge}
\acrodef{UWB}{Ultrawide bandwidth}
\acrodef{TNR}{threshold-to-noise ratio}
\acrodef{LS}{least squares}
\acrodef{IR-UWB}{impulse radio UWB}
\acrodef{FCC}{Federal Communications Commission}
\acrodef{TH}{time-hopping}
\acrodef{PPM}{pulse position modulation}
\acrodef{MUI}{multi-user interference}
\acrodef{PDP}{power delay profile}
\acrodef{BPZF}{band-pass zonal filter}
\acrodef{SIR}{signal-to-interference ratio}
\acrodef{SINR}{signal-to-interference-plus-noise ratio}
\acrodef{RFID}{radio frequency identification}
\acrodef{WPAN}{wireless personal area network}
\acrodef{WWB}{Weiss-Weinstein bound}
\acrodef{DP}{direct path}
\acrodef{MF}{matched filter}
\acrodef{MMSE}{minimum-mean-square-error}
\acrodef{SBS}{serial backward search}
\acrodef{SBSMC}{serial backward search for multiple clusters}
\acrodef{NBI}{narrowband interference}
\acrodef{WBI}{wideband interference}
\acrodef{INR}{interference-to-noise ratio}
\acrodef{CR}{channel response}
\acrodef{CIR}{channel impulse response}
\acrodef{CR}{channel  response}
\acrodef{RADAR}{radar}
\acrodef{MUR}{Multistatic radar}
\acrodef{JBSF}{jump back and search forward}
\acrodef{HDSA}{high-definition situation-aware}
\acrodef{RRC}{root raised cosine}
\acrodef{ST}{simple thresholding}
\acrodef{BTB}{Bellini-Tartara bound}
\acrodef{P-Max}{$P$-Max}  
\acrodef{MIMO}{multiple-input multiple-output}
\acrodef{MAP}{maximum a posteriori}
\acrodef{FG}{factor graph}
\acrodef{OP}{outage probability}
\acrodef{WED}{wall extra delay}
\acrodef{RMS}{root mean square}
\acrodef{SPAWN}{sum-product algorithm over a wireless network}
\acrodef{MDD}{minimum distance distribution}
\acrodef{MAP}{maximum a posteriori probability}
\acrodef{SAP}{small cell access point}
\acrodef{UE}{user equipment}
\acrodef{MBS}{macro cell base station}
\acrodef{UER}{\ac{UE} Relay}
\acrodef{D2D}{device-to-device}
\acrodef{MBS}{macro base station}
\acrodef{CSI}{channel state information}
\acrodef{OGR}{outage guard region}
\acrodef{FUR}{feasible UER region}
\acrodef{EHR}{energy harvesting region}
\acrodef{EH}{energy harvesting}
\acrodef{D2D-EHSN}{D2D communication provided \ac{EH} small cell network}
\acrodef{D2D-EHHN}{D2D communication provided \ac{EH} heterogeneous network}
\acrodef{3GPP}{3rd Generation Partnership Project}
\acrodef{BS}{base station}
\acrodef{DF}{decode and forward}
\acrodef{CCDF}{complementary cumulative distribution function}
\acrodef{ZF}{zero forcing}
\acrodef{RZF}{regularized zero forcing}
\acrodef{WLLN}{weak law of large number}
\acrodef{SLLN}{strong law of large numbers}
\acrodef{TDD}{Time-division duplex}
\acrodef{EE}{energy efficiency} 
\acrodef{HetNet}{heterogeneous network} 
\acrodef{SCP}{Single Cell Processing}
\acrodef{CBF}{Coordinated Beamforming}

%% file: defmetric.tex
\usepackage{color}
\usepackage{dsfont}
\usepackage{bbm}

\DeclareMathAlphabet{\mathsf}{OML}{cmbr}{m}{it}

\newtheorem{definition}{\bf Definition}
\newtheorem{theorem}{\bf Theorem}
\newtheorem{lemma}{\bf Lemma}

\newtheorem{remark}{\bf Remark}

\newtheorem{assumption}{\bf Assumption}

\newcommand{\bd}{\begin{description}}
\newcommand{\ed}{\end{description}}
\newcommand{\be}{\begin{enumerate}}
\newcommand{\ee}{\end{enumerate}}
\newcommand{\bi}{\begin{itemize}}
\newcommand{\ei}{\end{itemize}}
\newcommand{\bl}{\begin{list}}
\newcommand{\el}{\end{list}}
\newcommand{\bt}{\begin{tabbing}}
\newcommand{\et}{\end{tabbing}}